\documentclass[Afour,sageh,times]{sagej}

\usepackage{moreverb,url}
\usepackage{subcaption}
\usepackage[normalem]{ulem}
\usepackage{makecell}

\usepackage[colorlinks,bookmarksopen,bookmarksnumbered,citecolor=red,urlcolor=red]{hyperref}

\newcommand\BibTeX{{\rmfamily B\kern-.05em \textsc{i\kern-.025em b}\kern-.08em
		T\kern-.1667em\lower.7ex\hbox{E}\kern-.125emX}}

\begin{document}
	
	\runninghead{Zhang, Iida and Forni}
	
	\title{Periodic robust robotic rock chop via virtual model control}

	\author{Yi Zhang\affilnum{1}, Fumiya Iida\affilnum{2},  Fulvio Forni\affilnum{1}}
	
	\affiliation{\affilnum{1}University of Cambridge, United Kingdom. \affilnum{2}University of Tokyo, Japan.}
	
	\corrauth{Yi Zhang, University of Cambridge, United Kingdom.}
	
	\email{yz892@cam.ac.uk}
	
	\begin{abstract}
		
        Robotic cutting is a challenging, contact-rich manipulation task where the robot must simultaneously negotiate unknown object mechanics, large contact forces, and precise motion requirements. Our hypothesis is that this complexity can be alleviated through the design of a physically structured virtual-model controller that uses switched virtual mechanisms to generate a robust, rhythmic rock-chop motion for robotic cutting, without requiring pre-planned trajectories or precise environmental information. 
        Motion is generated by the interaction between the environment, the robot's dynamics, and the virtual forces of the switching virtual mechanism, ultimately realized through the available actuation. Through theoretical and numerical analysis, together with experimental validation, we demonstrate that the controlled robot behavior settles into a stable periodic motion. Experiments with a Franka manipulator demonstrate robust cuts across five different vegetables, achieving sub-millimeter slice accuracy for thicknesses from 1 mm to 7 mm at a rate of nearly one cut per second. The controller maintains high performance despite changes in knife shape or cutting board height, and successfully adapts to a different humanoid manipulator, demonstrating robustness and platform independence.
	\end{abstract}
	
	
	\maketitle
	
	\section{Introduction}
	
	Robotic cutting captures the complexity of contact-rich tasks. Cutting requires dexterous, precise motion that adapts to different surfaces, knife geometries, and food mechanics.
	Contact-rich manipulation is a central theme of robotics research, with significant progress achieved through learning-based approaches \cite{popov_data-efficient_2017, rajeswaran_learning_2018, gupta_reset-free_2021, radosavovic_state-only_2021, chen_system_2021} and model-based approaches \cite{posa_direct_2014, marcucci_approximate_2017, hogan_reactive_2020, aydinoglu_real-time_2022}. At the level of control algorithms, contact-rich interactions require impedance control \cite{Hogan1985a,Hogan1985b,Hogan1985c}.
	Advanced control over the force-displacement relationship at the end-effector is essential to regulate complex interactions with the environment. However, despite significant progress in robotic manipulation, autonomous cutting remains an open challenge due to the unpredictable nature of material fracture and deformation. In addition, the specific geometries of the knife and cutting surface are often only partially known. This uncertainty in the mechanical coupling between the tool and the workspace prevents the use of rigid, pre-programmed trajectories and requires a controller that can handle unmodeled geometric constraints.
	
	In the ‘trivial’ task of cutting in a kitchen, the knife repeatedly comes into contact with objects of varying material properties, ranging from soft fruits, such as strawberries, to stiffer objects, such as pumpkins and a rigid cutting board.
	The soft mechanics of food make accurate simulation challenging and this, in turn, complicates the application of learning- and optimization-based methods.
	In practical settings, learning approaches require a large number of interactions to converge to an effective control policy \cite{2015deepmpc,2019mitsioni,2021mitsioni,beltran-hernandez_sliceit_2024}, which not only produces significant food waste but also poses safety risks when experimenting with a knife. 
	\begin{figure}[t]
		\begin{center}
			\includegraphics[width=0.7\columnwidth]{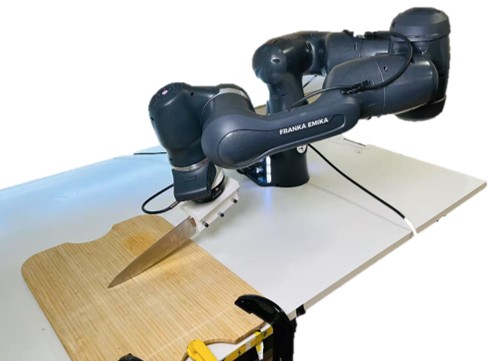} 
			\includegraphics[width=0.7\columnwidth]{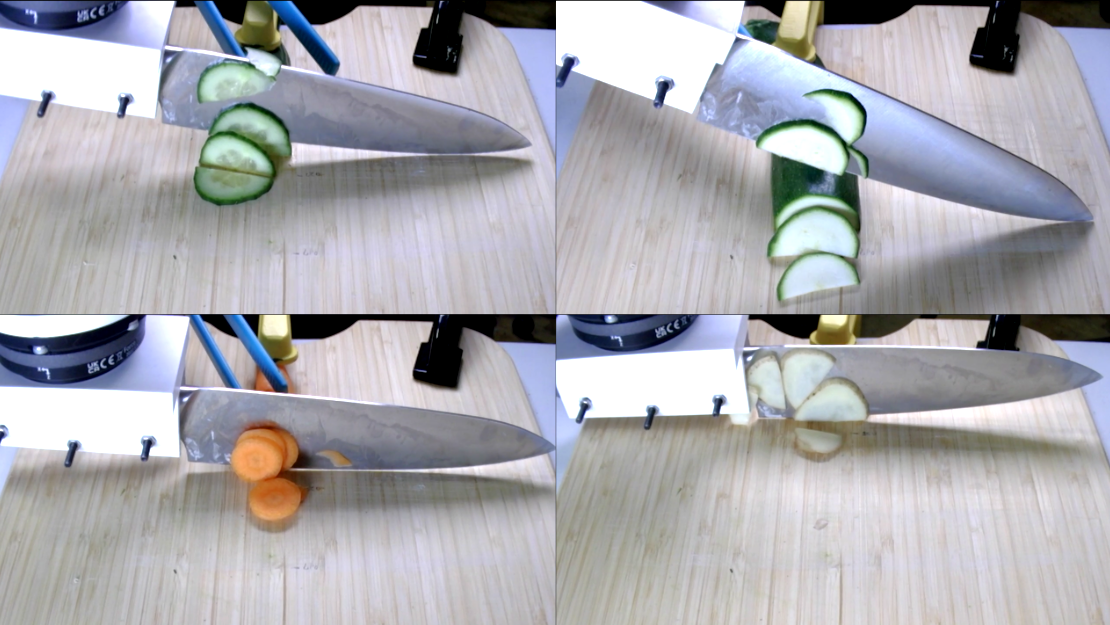}
			
			\caption{Robotic arm with knife rigidly attached, and snapshots of robotic cutting experiments. }
		\label{fig:slices_of_veges}
		\label{fig:experiment_setup}
	\end{center}
\end{figure}
Dynamic cutting techniques such as rocking chop are particularly challenging because the desired behavior is rhythmic. The rhythm must be produced while the blade repeatedly enters contact, fractures material, and rolls over a support surface.
Executing such motions via predefined trajectories requires precise kinematic models of the blade’s geometry \cite{2023YanBin}. Furthermore, it demands an exact characterization of the environmental constraints, specifically the height and planarity of the cutting board, as well as the evolving geometry of the food.
In contrast, it would be highly desirable to have a robust approach that does not depend on precise information, such as the knife geometry.
Conventional trajectory generation methods lack this generalization capability and typically rely on accurate vision sensing for re-planning \cite{2013Philippe,2014forcevision,Wright2024}.
Even if a policy is learned successfully in simulation, challenges remain when the robot encounters contact conditions, object properties, or tool geometries outside the training distribution. 
Increasing the amount of training data or domain randomization is one possible response to this problem. 
In this paper, we investigate a complementary direction: whether robustness can instead be obtained from the structure of the controller itself, so that the resulting behavior is shaped by physically meaningful interaction dynamics rather than by a precomputed trajectory or a learned mapping alone.

Impedance control \cite{Hogan1985a} is the standard paradigm for tasks involving stiff and uncertain contacts. However, identifying a reliable impedance for dynamic cutting remains a significant challenge. The system must be sufficiently stiff to penetrate the food, maintain a rectilinear cutting plane, and ensure accurate rolling kinematics. Conversely, it must remain compliant enough to guarantee reliable contact with the board, and moderate the contact forces and prevent jamming in the presence of geometric uncertainties. Whether these competing requirements can be reconciled with a static, uniform impedance remains an open question. Our experimental findings suggest that impedance must instead be modulated according to the specific geometry and phase of the cutting cycle. However, to the best of our knowledge, a formal theory for phase-dependent or adaptive impedance design in the context of robotic cutting with geometric uncertainties has yet to be established.

In this paper, we propose an approach that uses a physically structured controller to reduce dependence on accurate modelling information while enabling the generation of robust and complex rhythmic motions for contact-rich manipulation.
We deploy virtual model controller in a rhythmic contact-rich setting, where switched virtual mechanisms are used to regulate the cutting process into a limit cycle that produces self-sustained and robust rock-chop behavior. Our approach is rooted in the classical methodology of Virtual Model Control (VMC), originally developed to manage the stiff, complex interactions inherent in robotic locomotion \cite{Pratt2001VirtualMC, larby_optimal_2025}.
Rather than relying on pre-programmed trajectories, the desired behavior emerges from the dynamic interaction between the robot’s physical embodiment, the virtual mechanism, and the environment. Methodologically, our approach diverges from existing VMC literature by incorporating a regulated injection of energy, resulting in a stable, periodic rock chop motion. The limit cycle is achieved via controlled switching between two virtual mechanical topologies that correspond to the rising and falling phases of the knife.
Our goal is therefore not to solve the complete autonomous food-preparation pipeline, 
but to study how far a physically structured execution controller can go in producing robust, periodic cutting behavior under uncertainties.

Our design of the virtual mechanisms for cutting is based on intuitive physical choices, which can be described graphically in a simple way. Our virtual model controller does not require extensive training data, and parameter tuning is efficient. The controller modulates the robot impedance in a complex, nonlinear fashion throughout the cutting cycle, driven by real-time interaction features.
By leveraging its virtual components, the virtual mechanisms realize the necessary dynamics without requiring explicit trajectory or force tracking, and demonstrates robustness to environmental variations; the cutting action remains viable even if knife, workspace, or object geometry change, albeit with graceful performance degradation. Finally, the integration of sensory feedback enables online monitoring and optimization of the virtual parameters, facilitating the refinement of cutting frequency and force to ensure successful material separation.

The central finding of this paper is that virtual model control provides a robust design framework for complex manipulation, significantly reducing the reliance on high-fidelity modeling and the associated computational overhead. The specific contributions of this work are: 
\begin{itemize}
	\item A VMC-based controller with an original mechanism design that generates robust limit cycle for self-sustained rock chop cutting through controlled mechanical switching and regulated energy injection rather than explicit trajectory planning;
	\item Theoretical and numerical analyses, together with experimental validation, of the stability and periodicity of the cutting motion;
	\item Experimental characterization of cutting performance across various types of vegetables, for different thicknesses and cutting speeds;  
	\item Experimental validation of the controller’s robustness against perturbations of knife geometry and cutting board height;
	\item Demonstration of the controller’s portability (retargeting) through implementation on two distinct robotic platforms: the Franka FR3 and the RT Corp Sciurus17;
	\item A preliminary study of the online adaptation capabilities enabled by the proposed VMC design. 
\end{itemize}

The paper is organized as follows. A brief overview of the robotic cutting landscape is summarized in \textbf{Related works}. The methodology of virtual model control and a detailed description of the virtual model controller for cutting are provided in \textbf{Virtual cutting mechanisms design}. An energy-based analysis of the controlled robot behavior can be found in \textbf{Energy-based analysis}. \textbf{Experimental platform and cutting motion} provides the details of the implementation and an illustration of the robot motion. 
Experimental benchmarks are discussed in \textbf{Experimental result and discussion}, with a focus on performance and robustness.
Conclusions and further research directions are explored in \textbf{Conclusion and future works}, which also provides a detailed comparison with related works. 

\section{Related works}
\label{sec:related_works}
Predicting the mechanical interaction during food cutting is inherently challenging.
One of the reasons is that developing an accurate physical model of foodstuffs is complex. 
The force required to cut through a food item depends on numerous factors, including ripeness, skin presence, orientation, slice thickness, and the knife’s rake angle and sharpness \cite{Vincent_1991, 1993appleFracture, ATKINS200911, ATKINS2009283}. 
Simply applying a high cutting force can yield imprecise cuts or even damage delicate produce such as tomatoes. 
In earlier studies, cutting force was often treated as an external disturbance, with adaptive controllers employing position and velocity feedback to maintain predetermined trajectories \cite{1997adaptive}. 
Impedance control was introduced to regulate cutting force directly, augmented by adaptive laws to accommodate time-varying desired positions and environments \cite{1999impedance}.

Robotic cutting is commonly approached as a motion tracking problem, often informed by visual sensing. 
A key challenge lies in formulating the cutting trajectory itself. While basic trajectories can be designed, more sophisticated cutting trajectories are harder to derive.
Remarkably, 
\cite{2013Philippe} obtained a smoothed path from camera data to separate meat muscle.
\cite{2014forcevision} proposed a cut and slice trajectory adjusted through force sensing, to reduce damage to deformable objects. 
More complex rolling (or rocking) actions have been addressed by leveraging the knife geometry to maintain a constant contact force \cite{2023YanBin}. 
\cite{2023Ramamoorthy} explored learning from demonstration, introducing a coupling term via the Udwadia–Kalaba method to suppress lateral blade movements.
Reinforcement learning is used in \cite{beltran-hernandez_sliceit_2024} to learn a cutting trajectory and impedance parameters.
Because developing an exact physical model of foodstuffs is challenging, learning-based approaches have also garnered attention. A series of works \cite{2015deepmpc, 2019mitsioni, 2021mitsioni} employs data-driven model predictive control (MPC), where a neural network is used to predict future knife positions based on current position and force measurements, thus enabling real-time optimization of the reference force. 
In \cite{2020Inna}, variational autoencoders are used to model the cutting dynamics, though their primary goal is to detect material properties and thickness rather than to learn a cutting strategy.

The main difference with our approach is that our controller can generate complex rocking motion without requiring a predefined trajectory. Our controller design and deployment require minimal information about the food and the external environment (knife geometry, cutting board features). Furthermore, its design is not data-intensive and both design and deployment require minimal computing efforts. In turn, these features guarantee the robot achieves successful cutting, robustly handling different food types.

A second challenge involves managing unanticipated interactions with the environment. 
Inaccurate camera-based detection of the object position, or variations in the table height, can lead to collisions or excessive forces if the robot strictly follows a fixed cutting path. 
This often results in safety triggers when the commanded position is physically unachievable (e.g., below the cutting board). 
To deal with the interaction,  \cite{2019YanBin,2023YanBin} adopt a hybrid scheme that switches to impedance or direct force control upon detecting contact with the cutting surface. 
Their method segments the cutting process into distinct press, push, and slice phases, applying different controllers in each stage.
In our approach, the controller does not require the identification of different stages of cutting. Rocking is the result of the continuous interaction between robot mechanics, virtual model controller, and environment. Unanticipated interactions may degrade the performance,
but only large perturbations will disrupt the whole rocking motion.

A consistent benchmarking framework remains largely absent in the field of robotic cutting. 
Although prior work often reports end-effector trajectory and force sensor data for individual cutting trials, these metrics appear insufficient for real-world applications. 
Reliability, that is, the ability to repeatedly produce separated slices of accurate thickness and handle variations in the environment (e.g., knife, board height, or food properties), remains understudied. In contrast, in this paper we quantify the performance of our controller against a broad set of relevant metrics, such as cut frequency, cutting force, cutting speed, slice-thickness accuracy and variance, and we demonstrate successful cuts across different environmental conditions.

\section{Virtual cutting mechanisms design} \label{methodology}

\subsection{Virtual model control in a nutshell}

A virtual model controller consists of a (virtual) mechanical linkage combined with springs, dampers, and masses that, once coupled with the robot, shapes its behavior. It achieves this by imposing constraints on the motion of the robot and its reaction to the environment. The associated virtual forces are then realized through the robot's available actuation, typically mapped into joint torques and executed by the motors at each joint \cite{Pratt2001VirtualMC, 10801592, larby_optimal_2025}.
This mapping relies on the robot's kinematics.

For fully actuated robot manipulator based on revolute joints, using $\pmb{q}$ to denote the joint coordinates of the robot, the force $\pmb{f}_i$ generated by a virtual component at a point $\pmb{p}_i = \pmb{h}_i(\pmb{q})$ of the robot is mapped into corresponding joint torques $\pmb{\tau}$ using the Jacobian matrix $J_i(\pmb{q}) = \frac{\partial \pmb{h}_i(\pmb{q})}{\partial \pmb{q}}$.
From the principle of virtual work \cite{Spong2005}, the torque command to emulate the forces of the virtual model controller is given by 
\begin{equation}
	\label{eq:translation}
	\pmb{\tau} = J_1(\pmb{q})^T \pmb{f}_1 + J_2(\pmb{q})^T\pmb{f}_2 + J_3(\pmb{q})^T\pmb{f}_3 + \dots
\end{equation}
Virtual model control is intrinsically stable, because the controlled robot is the interconnection of the passive mechanics of the robot and the passive virtual mechanics \cite{larby_optimal_2025}.
Stability is guaranteed by passivity theory \cite{Willems1972,VanDerSchaft1999,Stramigioli2007,Stramigioli2015,Stramigioli2017,Stramigioli2020,Ortega2023pid}. 
Virtual model control provides an intuitive, task-oriented method for energy shaping and damping injection in robotics \cite{Ortega2001,Ortega1998}. 

\begin{figure}[htbp]
	\centering
	\includegraphics[width=.4\columnwidth]{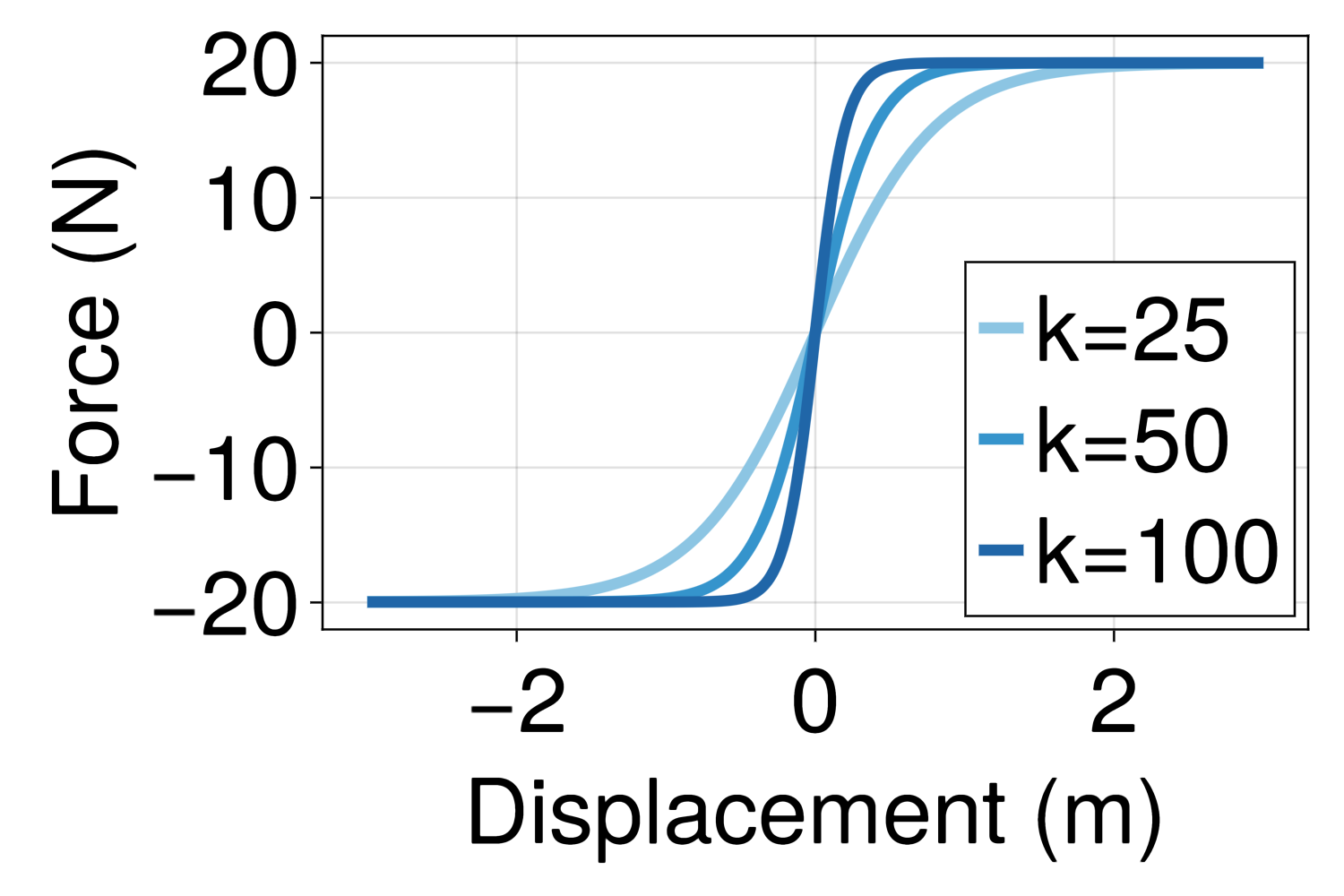}
	\vspace{-1mm}
	\caption{$20$ N saturated nonlinear spring force profile.}
	\label{fig:nonlinear_spring}
\end{figure}

The selection of spring and damper parameters and their geometric configuration also shape the impedance of the robot \cite{Hogan1985a,Colgate1988}.
Although impedance control is typically implemented with linear springs and dampers, nonlinear springs featuring force saturation can limit excessively rapid movements and enhance safety.
In this work, all springs in the virtual mechanism have a saturating nonlinear characteristic 
between displacement and force given by
\begin{equation}
	\pmb{f}(k, \sigma, \pmb{z}) = \sigma \tanh(k|\pmb{z}|/\sigma)\frac{\pmb{z}}{|\pmb{z}|},
\end{equation}
where $\pmb{z} \in \mathbb{R}^3$ represents the spring displacement, $k$ is the stiffness parameter, the maximum force is governed by $\sigma\geq 0$, 
with $\pmb{f}$ assumed to be 0 at $|\pmb{z}|=0$.
Fig. \ref{fig:nonlinear_spring} shows the force profile of a spring with stiffness 25 N/m and maximum force 20 N.

\subsection{Virtual model control for cutting}

\begin{figure*}
	\centering
	
	\begin{subfigure}[t]{\textwidth}
		\centering
		\includegraphics[width=0.7\linewidth]{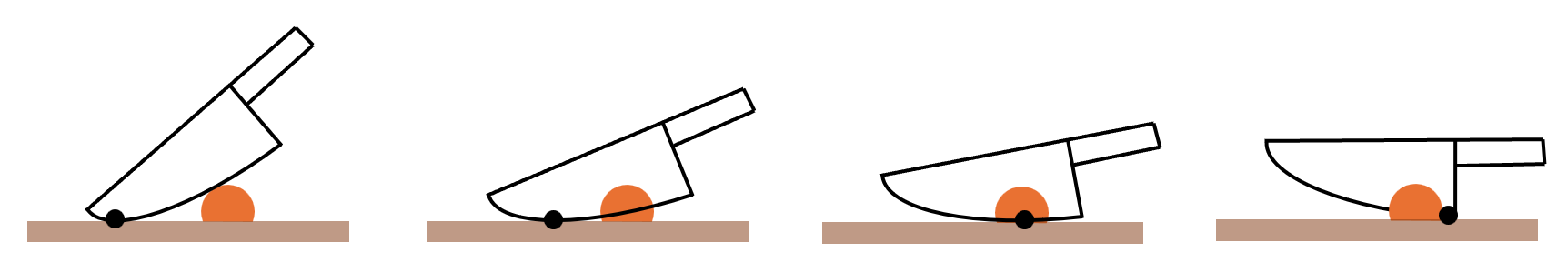}
		\caption{\textbf{Rocking motion overview.} The knife pivots from tip to heel against the cutting board.}
		\label{fig:rocking_illustration}
	\end{subfigure}
	\vspace*{0.3cm}
	
	\begin{subfigure}[t]{\textwidth}
		\centering
		\includegraphics[width=0.7\linewidth]{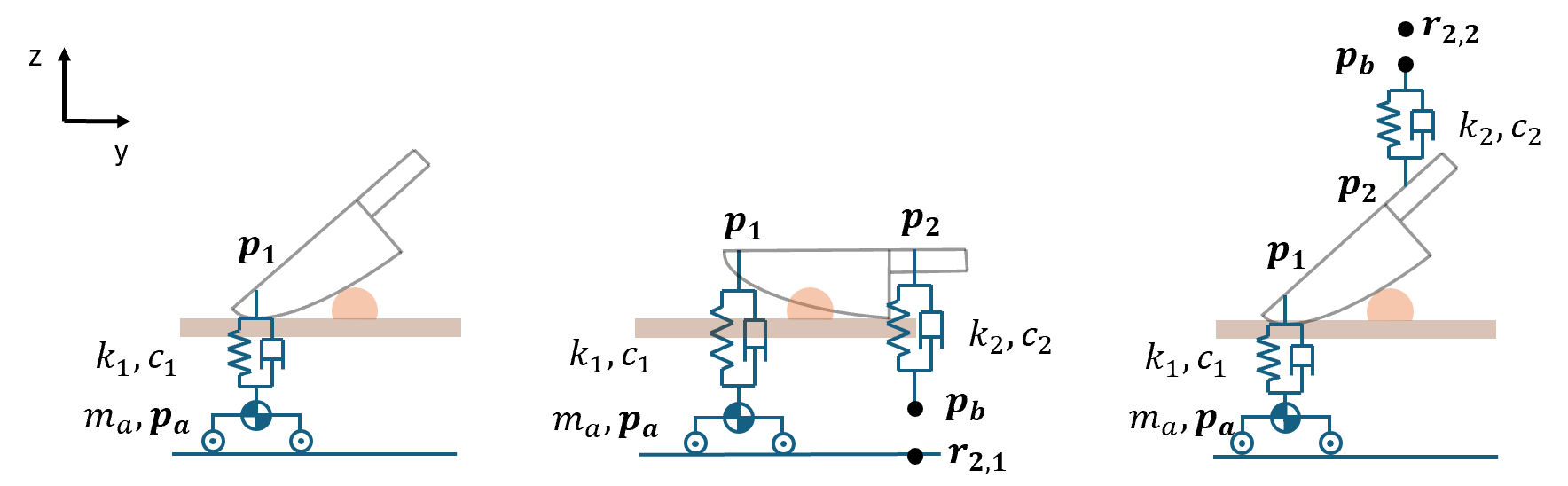}
		\caption{\textbf{Left: Maintaining contact.} A spring and damper link the knife tip $\pmb{p}_1$ to $\pmb{p}_a$, keeping the blade in contact with the board while being able to slide along the rocking direction ($y$ axis). \textbf{Middle and right: Cutting and raising.} Spring-damper pair connects the knife handle $\pmb{p}_2$ to $\pmb{p}_{b}$ which moves vertically for cutting and raising the knife. When switch from cutting to raising, the reference points are shifted along $x$ axis by one slice thickness to produce multiple slices.}
		\label{fig:rocking_only}
	\end{subfigure}
	\begin{subfigure}[t]{\textwidth}
		\centering
		\includegraphics[width=0.5\linewidth]{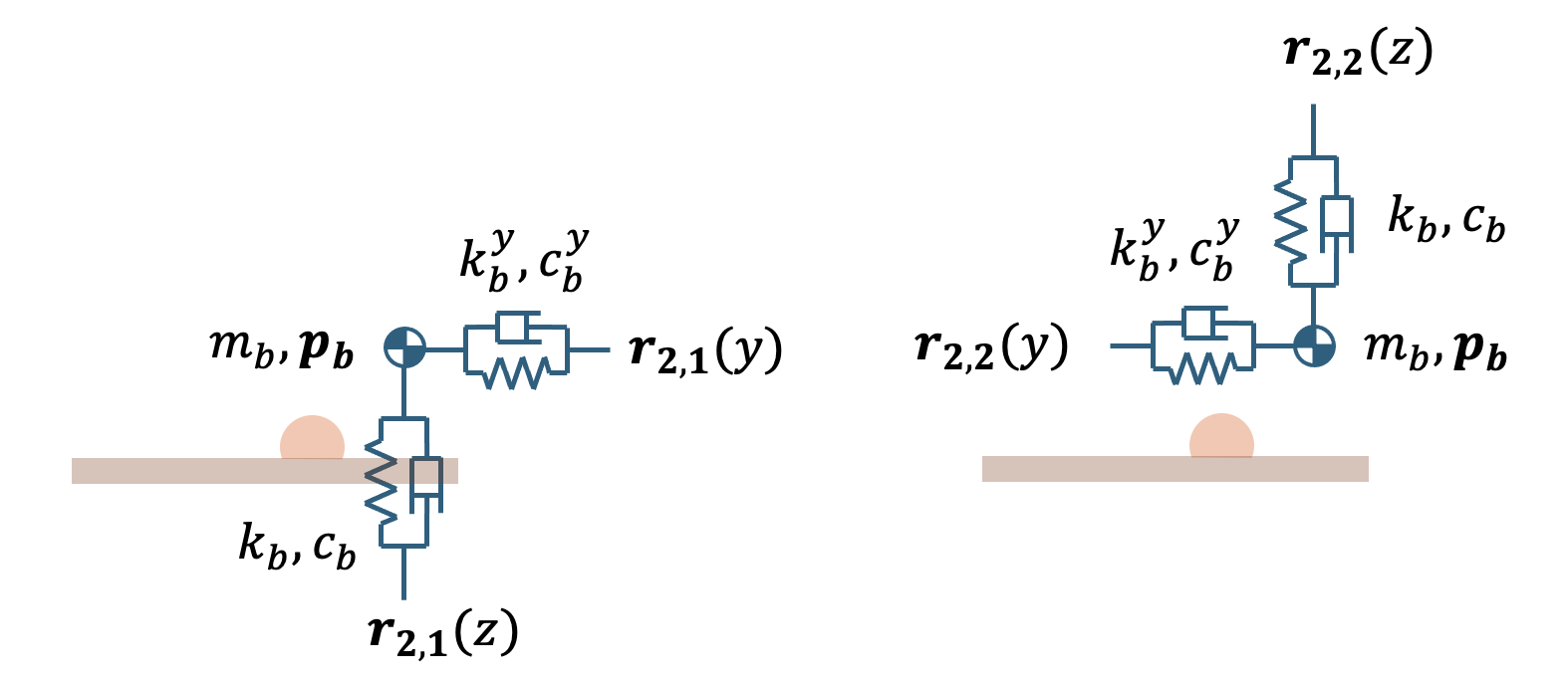}
		\caption{\textbf{Switching references smoothly for cutting (left) or raising (right).} Spring-damper pair connects $\pmb{p}_b$ to $\pmb{r}_{2,1}$ for cutting. Switching from $\pmb{r}_{2,1}$ to $\pmb{r}_{2,2}$ raises the knife. Instead of directly connecting $\pmb{p}_2$ to $\pmb{r}_{2,1}$ or $\pmb{r}_{2,2}$, the mass-spring-damper here produces a smoother transition. 
        $\pmb{p}_b$ slides in the $y-z$ plane. $k_b^y$ is set high enough so the transition in $y$ axis is fast to overcome the friction due to cutting force. 
        }
		\label{fig:rocking_only_smooth}
	\end{subfigure}
	\begin{subfigure}[t]{\textwidth}
		\centering
		\includegraphics[width=0.7\linewidth]{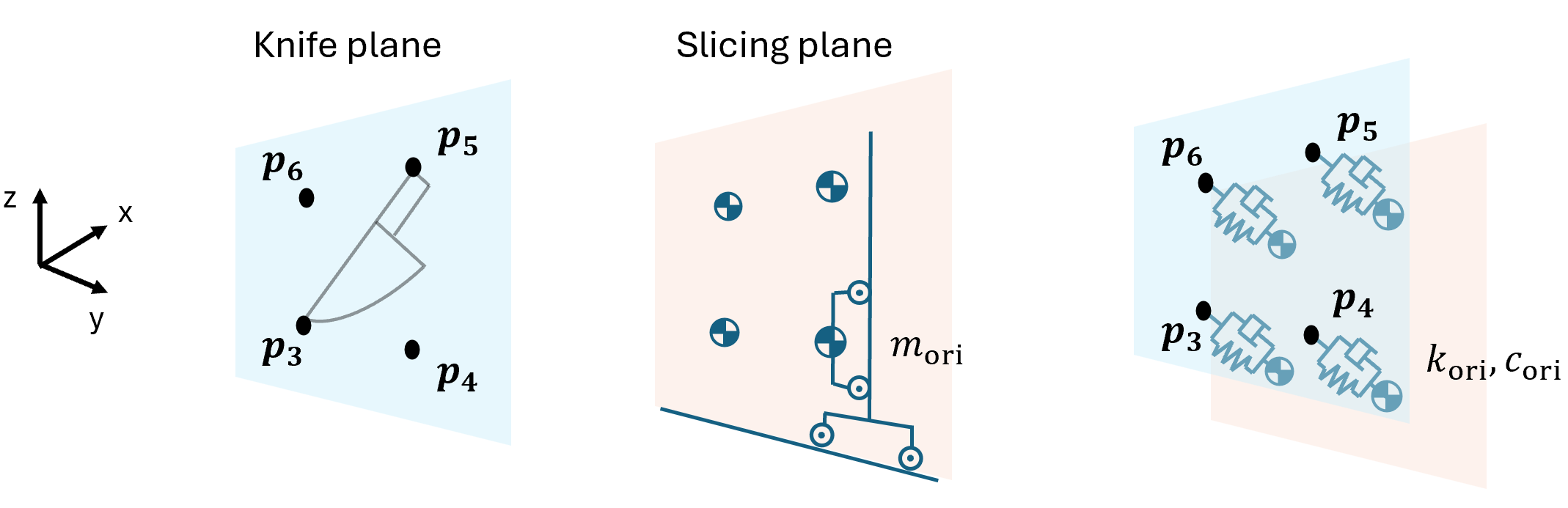}
		\caption{
			\textbf{Constraining the slicing plane.} Left: Define four fixed points $p_3,\dots,p_6$ relative to the knife. Middle: Define four virtual masses, each can slide along the two axes in the desired slicing plane. Right: Each of the points $p_3,\dots,p_6$ is connected to an individual point mass with a spring and damper.
		}
		\label{fig:rocking_orientation}
	\end{subfigure}
	
	\caption{Virtual mechanism design for rocking motion.}
	\label{fig:rocking_mechanism}
\end{figure*}

A rocking motion involves continuously shifting the knife’s contact point along the cutting board, from its tip to its heel, as illustrated in Fig. \ref{fig:rocking_illustration}.
To achieve this using a virtual mechanism, we first define a virtual mass at
$\pmb{p}_a$ that is constrained to move only along the rocking direction ($y$ axis in Fig. \ref{fig:rocking_only}). Its $x$ coordinate is fixed to
the cutting plane and its $z$ coordinate is fixed at a height which is under the cutting-board, while its $y$ coordinate is free to evolve. A spring and damper connect the knife tip at $\pmb{p}_1$ to the virtual mass at $\pmb{p}_a$, ensuring that the blade remains in contact with the surface.
The goal of the free-sliding virtual mass is to ensure that the knife can move freely while keeping contact with the board, without interfering with the rocking motion.
Since the virtual mass at $\pmb{p}_a$ has only one degree of freedom, its motion is governed by the $y$-component of the spring damper interaction:
{\color{black}
\begin{align}
m_a\ddot{y}_{\pmb{p}_a}
+ \pmb{e}_y^T
\pmb{f}(k_1,\sigma_1,\pmb{p}_a-\pmb{p}_1)
+ c_1\left(
\dot{y}_{\pmb{p}_a}-\dot{y}_{\pmb{p}_1}
\right)=0
\end{align}
where $\pmb{e}_y=[0,1,0]^T$ selects the $y$-component
of the three-dimensional spring force,
$y_{\pmb{p}_{1}}$ is the position of $\pmb{p}_{1}$ in the slicing direction, and $c_1$ is the damping coefficient.
}

Another spring and damper connect the knife handle at $\pmb{p}_2$ to the control point $\pmb{p}_{b}$ for the cutting motion. 
Switching $\pmb{p}_{b}$ from reference $\pmb{r}_{2,1}$ to reference $\pmb{r}_{2,2}$ raises the knife. Switching from $\pmb{r}_{2,2}$ to $\pmb{r}_{2,1}$ cuts down (Fig. \ref{fig:rocking_only}). 
Hence, the system alternates between a cutting state and a knife-raising state, producing the rocking action.
The motion is in cutting phase when $\pmb{p}_b$ is connected to $\pmb{r}_{2,1}$ and raising phase when connected to $\pmb{r}_{2,2}$.

Positions $\pmb{p}_1$ and $\pmb{p}_2$ are defined relative to the robot's end-effector, assuming that the knife grip is rigid enough, so that the relationship between end-effector and knife does not change.
$\pmb{p}_1$ and $\pmb{p}_2$ do not have to lie on the physical knife. 
However, their relative geometry should preserve the following conditions:
$\pmb{p}_1$ is towards the direction of knife tip and $\pmb{p}_2$ is in the direction of handle. The straight line connecting $\pmb{p}_1$ to $\pmb{p}_2$ should be approximately aligned with the knife spine. 
Accordingly, the condition $z_{\pmb{p}_1}\approx z_{\pmb{p}_2}$ indicates that the knife spine is approximately aligned with the cutting board, which is used as the criterion for completion of the cutting phase.
$\pmb{p}_a$, $\pmb{r}_{2,1}$ and $\pmb{r}_{2,2}$ lie in the world frame and depend on the position of the food.
The difference in $y$ coordinate of $\pmb{r}_{2,1}$ and $\pmb{r}_{2,2}$ regulates the slicing distance.
The cutting force is regulated through the stiffness parameters $k_1$ and $k_2$.

The switch from cutting to raising is triggered by the knife's orientation. 
When $z$ coordinates of $\pmb{p}_1$ and $\pmb{p}_2$ are close (within tolerance $\delta_1$), the knife spine (the line that passes $\pmb{p}_1$ and $\pmb{p}_2$) aligns with the cutting board surface and the cut is done.
The switch from raising to cutting is triggered when $\pmb{p}_2$ is close to $\pmb{r}_{2,2}$ (within tolerance $\delta_2$ measured along the z coordinates).

\begin{subequations}
	All together, these components produce virtual forces
	\begin{align}
		\pmb{f}_1 &= \pmb{f}(k_1, \sigma_1, \pmb{p}_a-\pmb{p}_1)+c_1(\dot{\pmb{p}}_a-\dot{\pmb{p}}_1) \label{eq:simple_rocking} \\
		\pmb{f}_2 &= \pmb{f}(k_2, \sigma_2, \pmb{p}_b-\pmb{p}_2)+c_2(\dot{\pmb{p}}_b-\dot{\pmb{p}}_2).
	\end{align}
	where $\pmb{p}_b$ is switched between $\pmb{r}_{2,1}$ and $\pmb{r}_{2,2}$ depending on whether it is in the cutting or raising phase. The way $\pmb{p}_b$ switches is also modulated by the dynamics in Fig. \ref{fig:rocking_only_smooth}. 
	We smooth the transition by placing a mass that slides along the line between $\pmb{r}_{2,1}$ and $\pmb{r}_{2,2}$. The mass is rigidly attached to $\pmb{p}_b$ and connected to $\pmb{r}_{2}$ by spring and dampers.
\end{subequations}

Finally, to ensure the knife moves primarily within the plane that defines the slice, we introduce four additional points $\pmb{p}_3, \dots, \pmb{p}_6$ on the knife (Fig. \ref{fig:rocking_orientation}) and connect each to a virtual mass free to slide along the desired slicing plane, whose position
is denoted by the reference $\pmb{r}_S$.
Each pair is linked by its own spring and damper.
This virtual mechanism enforces a plane-constrained motion, yielding a cleaner cut.
Attachment points $\pmb{p}_3$–$\pmb{p}_6$ are defined relative to the knife plane with respect to the end effector. These points need not lie on the knife itself; increasing their distance from the blade enhances orientation correction. While three points are theoretically sufficient to constrain orientation, using four provides improved stability without further increasing stiffness.
To produce multiple slices in sequence, once a slice is completed and the knife has returned to its raised position, we shift the slicing plane $\pmb{r}_S$, $\pmb{p}_a$, $\pmb{r}_{2,1}$, and $\pmb{r}_{2,2}$ to the next desired location.
Specifically, for slice $i$, the references are translated along the $x$ direction by the slice thickness $\Delta$:
\begin{align}
    \pmb{r}_S^{i+1} &= \pmb{r}_S^{i} + [1,0,0]\Delta.
\end{align}
The same update is applied to the $x$ coordinates of $\pmb{p}_a$, $\pmb{r}_{2,1}$, and $\pmb{r}_{2,2}$, while their $y$ and $z$ coordinates remain unchanged.

\subsection{Cutting performance metrics}
To quantitatively evaluate the cutting process, we have used the following metrics. Let $\mathcal{C}_{\mathrm{cut},i} = \{t_1,\dots,t_{N_i}\}$ 
denote the set of sampling times during the cutting phase $i$. These belong to the time intervals in which
$\pmb{p}_b$ is connected to the reference $\pmb{r}_{2,1}$, as shown in Fig. \ref{fig:rocking_only_smooth} (left).
The average and peak forces during the $i$th cycle are computed from the 
force-torque sensor reading, mounted at the robot wrist
\begin{equation}
	F_{\text{avg},i} 
	\,= \!\!
	\sum_{t\,\in\, \mathcal{C}_{\mathrm{cut},i}} \!\!\frac{|\pmb{F}(t)|}{N_i}
	\qquad\quad
	F_{\text{peak},i}
	\,=
	\max_{t \,\in\, \mathcal{C}_{\mathrm{cut},i}} |\pmb{F}(t)|.
	\label{eq:force_computation}
\end{equation}
Similarly, the average and maximum velocities of the robot wrist during cutting are estimated as follows.
\begin{equation}
	v_{\text{avg},i} 
	\,= \!\!
	\sum_{t\,\in\, \mathcal{C}_{\mathrm{cut},i}} \!\!\frac{|\pmb{v}(t)|}{N_i}
	\qquad\quad
	v_{\text{peak},i}
	\,=
	\max_{t \,\in\, \mathcal{C}_{\mathrm{cut},i}} |\pmb{v}(t)|,
\end{equation}
where $\pmb{v}$ is the velocity of the wrist, computed using the robot kinematics.
Finally, cut frequency is defined as
\begin{equation}
	f_{\text{cut}}
	\,=\,
	\frac{N_{\text{switch}}-1}{T}, 
\end{equation}
where $N_{\text{switch}}$ is the total number of transitions from the cut to the raise state that occur during the time interval $T$ of a whole experiment. This frequency indicates how rapidly consecutive slices are produced.

\subsection{Parameter selection and tuning guidelines}
	The proposed controller contains several parameters, but they have distinct physical roles. 
    In practice, tuning is performed by first selecting the geometric configuration of the virtual mechanism and then adjusting a smaller set of force and timing parameters. 

	\emph{Tuning of the mechanism geometric configuration:}
	The points $p_1$ and $p_2$ define the effective knife tip and handle directions used by the rocking mechanism. These points do not need to coincide exactly with physical points on the blade; approximate placement is sufficient. The cutting reference $\pmb{r}_{2,1}$ must be under the cutting board though the accurate position is not important. 
    The raising reference $\pmb{r}_{2,2}$ accounts for the food height and determines how far the knife is lifted before the next cut.
    If $\pmb{r}_{2,2}$ is set too low, the conditions for switching from raising to cutting and from cutting to raising can be satisfied simultaneously. This can prevent the controller from producing effective cutting, as observed in the only failure case in Table \ref{tab:sensitivity}. The horizontal displacement between $\pmb{r}_{2,1}$ and $\pmb{r}_{2,2}$ sets the slicing distance. The desired cutting plane is specified by the slicing-plane reference $\pmb{r}_S$. The parameters $\pmb{p}_3$–$\pmb{p}_6$, $k_{\mathrm{ori}}$, $\sigma_{\mathrm{ori}}$, and $c_{\mathrm{ori}}$ define the virtual springs and dampers that constrain the knife to the slicing plane. Points of $\pmb{p}_3$–$\pmb{p}_6$ need not lie exactly on the knife. Increasing $k_{\mathrm{ori}}$ or distance from $\pmb{p}_3$–$\pmb{p}_6$ to the blade improves slice evenness and suppresses lateral blade motion. 
	
	\emph{Contact and cutting force tuning:}
	The parameters $k_1$, $\sigma_1$, and $c_1$ determine how strongly the knife is pulled toward the cutting board through the spring-damper pair connected to $p_1$. Increasing $k_1$ or $\sigma_1$ improves contact stability. The parameters $k_2$, $\sigma_2$, and $c_2$ determine the force generated by the handle-side mechanism connected to $p_2$. Increasing $k_2$ or $\sigma_2$ increases the cutting force and can improve separation of harder objects, but may also increase peak contact force and residual energy after a cut. 
	The parameters $m_b$, $k_b$, and $c_b$ define the mass-spring-damper system that moves the intermediate point $p_b$ between $\pmb{r}_{2,1}$ and $\pmb{r}_{2,2}$. These parameters are lumped into a simple time constant $\tau$ in Table \ref{tab:sensitivity}. A faster $\tau$ increases the cut frequency and contact force. 
	
	\emph{Switching thresholds:}
	The thresholds $\delta_1$ and $\delta_2$ determine when the controller switches between cutting and raising. 
    Larger thresholds are useful at higher speeds because they prevent excessive rolling or raising. For smaller objects, $\delta_2$ can be increased to reduce the lifting height and allow higher cutting frequencies. 
	
	\emph{Porting to different robot platforms:}
	When transferring the controller to a different robot, the virtual mechanism geometry can be kept unchanged, while force-related parameters should be scaled according to the actuation capability, communication rate, and maximum payload of the platform. 
	
	Overall, the tuning procedure is:
	\begin{enumerate}
		\item Choose $p_1$, $p_2$, $\pmb{r}_{2,1}$, $\pmb{r}_{2,2}$, and $\pmb{r}_S$ from the approximated knife pose, food position, cutting board height and desired slice thickness. Select $\pmb{p}_3$–$\pmb{p}_6$, $k_{\mathrm{ori}}$, $\sigma_{\mathrm{ori}}$, and $c_{\mathrm{ori}}$ to maintain the slicing plane.
		\item Set $k_1$, $k_2$, $\sigma_1$ and $\sigma_2$ so that food is successfully separated while contact with the cutting board is maintained during rolling. Adjust $m_b$, $k_b$, and $c_b$ for the desired cut frequency.
		\item Tune $\delta_1$ and $\delta_2$ to reduce overshoot and excessive contact forces, thus mitigating wear on the knife and cutting board.
    \end{enumerate}

    The functional roles of the different parameters and their effects on system behavior are supported by the set of simulations in Table \ref{tab:sensitivity}, and later confirmed experimentally. The sensitivity analysis summarized in Table \ref{tab:sensitivity} is based on MuJoCo simulations using models of the Franka FR3 manipulator, a cutting board, and a knife, configured to match the physical experimental setup in later sections as closely as possible. Most \emph{nominal} parameter values are shared between the simulations and experiments, as shown in Tables \ref{tab:sensitivity_parameters} and \ref{tab:robust_experiment}. A few parameters were adjusted to account for the intrinsic gap between the simulated dynamics and the physical system.

    For simplicity, in the simulations the virtual mechanism was extended with an additional virtual spring and damper attached between the elbow and a fixed point in Cartesian space. This (weak) coupling had the effect of removing the null-space of the robot, 
    preventing the elbow from drifting. 
    To evaluate the sensitivity of the robot behavior to parameter \emph{perturbations}, we varied one virtual-mechanism parameter at a time by $\pm 20\%$, while keeping all other parameters fixed. For each successful case, we extracted the cutting performance metrics. The full numerical results are reported in Table \ref{tab:sensitivity}.
    Notably, we had a single failure case, due to insufficient separation between switching constraints ($\delta_1$ and $\delta_2$), causing fast switching between the two reference points.
    
    The results in Table \ref{tab:sensitivity} indicate that the controller is robust to parameter uncertainty. Variations in point placement, stiffness, damping, and switching thresholds generally change the cutting force, cut frequency, or motion profile, rather than making the rocking motion infeasible.

\section{Theoretical analysis of the cutting motion}

\subsection{Energy-based analysis}
\label{sec:energy}

Energy considerations provide insight into the emerging periodic motion.
The mechanical energy of the robot, $E_R$, satisfies the classical passivity inequality \cite{spong2022_annual_review, Spong2005, Ortega1998}
\begin{equation}
	\label{eq:E_R}
	\dot{E}_{\mathrm{R}} \leq \dot{\pmb{q}}^T \pmb{\tau} + \sum_i \dot{\pmb{s}}_{i}^T \pmb{d}_i
\end{equation}
where $\pmb{q}$ is the vector of joint coordinates of the robot manipulator, $\pmb{\tau}$ are the related motor torques, and $(\pmb{s}_{i},\pmb{d}_i)$ are additional co-located positions/forces variables, accounting for generic external actions to the robot due to the interaction of the knife with food and cutting board.
The inequality takes into account unavoidable losses due to internal mechanical dissipation.

The controller is also a (virtual) mechanism, whose mechanical (virtual) energy satisfies 
\begin{equation}
	\label{eq:E_VMC}
	\dot{E}_{\mathrm{VMC}} = -\sum_ {i=1}^6 \dot{\pmb{p}}_i^T \pmb{f}_i - W_{\mathrm{dissipation}} + W_{\mathrm{elastic}}.
\end{equation}
$W_{\mathrm{dissipation}} \geq 0$
accounts for the power dissipated in the dampers of the virtual mechanism. $W_{\mathrm{elastic}}$ captures
the power flow caused by the switching of the references $\pmb{r}_{2,1}$, $\pmb{r}_{2,2}$, and $\pmb{r}_{S}$.
The negative sign of the term 
$-\sum_ {i=1}^6 \dot{\pmb{p}}_i \pmb{f}_i$
in \eqref{eq:E_VMC} is due to the mechanical interconnection between robot and virtual model controller. Through \eqref{eq:translation}, each force $\pmb{f}_i$ acting on the robot exerts an equal and opposite force on the virtual mechanism. 

Taking into account the identity
$
\dot{\pmb{q}}^T \pmb{\tau}
= 
\dot{\pmb{q}}^T \sum_{i=1}^6 J_i(\pmb{q})^T \pmb{f}_i
=
\sum_{i=1}^6 \dot{\pmb{p}}_i^T \pmb{f}_i
$, 
the total 
energy of the controlled robot
$E = E_{\mathrm{R}} + E_{\mathrm{VMC}}$ satisfies
\begin{equation}
	\label{eq:total_energy}
	\dot{E} \leq - W_{\mathrm{dissipation}} + W_{\mathrm{elastic}} 
	+
	\sum_i \dot{\pmb{s}}_{i}^T \pmb{d}_i
\end{equation}
\eqref{eq:total_energy} shows that the energy of the controlled robot must continuously decrease if no external forces act on the robot (i.e. $\pmb{d}_i = 0$) and if we do not switch references $\pmb{r}_{2,1}$, $\pmb{r}_{2,2}$, and $\pmb{r}_{S}$ of the virtual mechanism.
Likewise, the energy necessarily decreases also if the external forces $\pmb{d}_i$ represent interaction forces with the food and cutting board. This follows from passivity theory
\cite{Willems1972,Ortega1998,VanDerSchaft1999,Stramigioli2007}, as the complete mechanical system consisting of the controlled robot interacting with food and cutting board can be modelled as the negative feedback interconnection of passive systems. 
In both scenarios, without switching, the robot motion eventually settles in a resting position. For the gravity-compensated robot, this corresponds to a minimum of the elastic potential energy. 
This is a classical result of passivity-based control \cite{Ortega2001}.

In our case, each time a reference is switched at a specific robot configuration, a bounded amount of energy is injected into or removed from the system. This action continuously drives the controlled robot into motion. The switch  between the cutting and raising phases is governed by specific triggers: the transition from raising the knife
to cutting is based on the position of the knife, while the switch from cutting to raising is mainly triggered by the orientation of the knife. When a switch occurs, the spring connecting $\pmb{p_b}$ to $\pmb{r}_{2,i}$, $i \in \{1,2\}$, is instantly detached from $\pmb{r}_{2,1}$ and attached to $\pmb{r}_{2,2}$ and vice-versa. 
The injected/removed energy corresponds to the discontinuity in elastic potential energy resulting from the instantaneous change in spring extension. 

The periodic cutting motions of the controlled robot, illustrated in the experimental results section below, emerge at the equilibrium between the energy injected via reference switching and the total dissipation from virtual damping, robot internal dissipation, and environmental resistance (i.e. food and cutting board).
Along any periodic motions of period $\bar{T}$ the controlled robot must satisfy 
\begin{equation}
	\int_t^{t+\bar{T}} \!\!\!\!\!\!\! W_{\mathrm{elastic}}(\tau) d\tau =\! 
	\int_t^{t+\bar{T}} \!\!\!\!\!\!\! W_{\mathrm{dissipation}}(\tau) \!+\! 
	\sum_i \dot{\pmb{s}}_{i}^T(\tau) \pmb{d}_i(\tau)  d\tau
\end{equation}
for all times $t$ (where we have neglected the internal dissipation of the robot, for simplicity).

The energy injected at each switch is dissipated by the robot's inherent damping and the virtual dampers, or transferred to the environment through interaction with the food and the cutting board. Any energy surplus ultimately manifests as increased velocity or higher cutting forces, both of which enhance dissipation. Due to the geometry of the compliant constraints imposed on the knife by the virtual mechanisms, the switch-induced variations primarily affect the knife’s planar velocity and its contact forces. Hence, over multiple cycles, the system settles into a stable periodic regime.

\begin{figure}[htbp]
	\centering
	\includegraphics[width=.4\columnwidth]{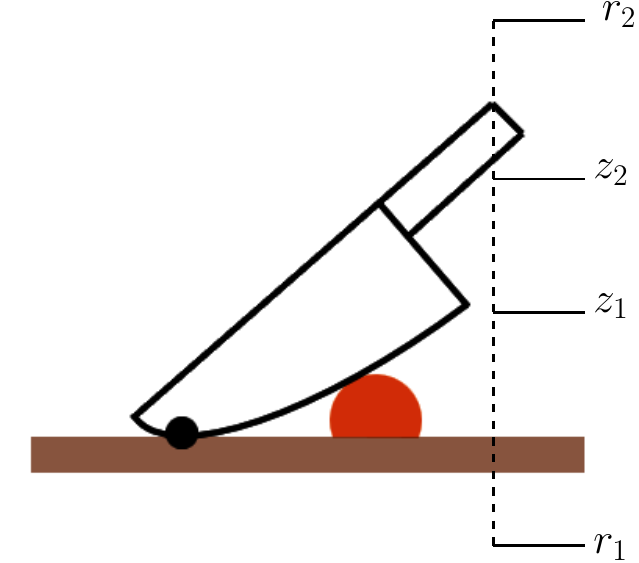}
	\includegraphics[width=.7\columnwidth]{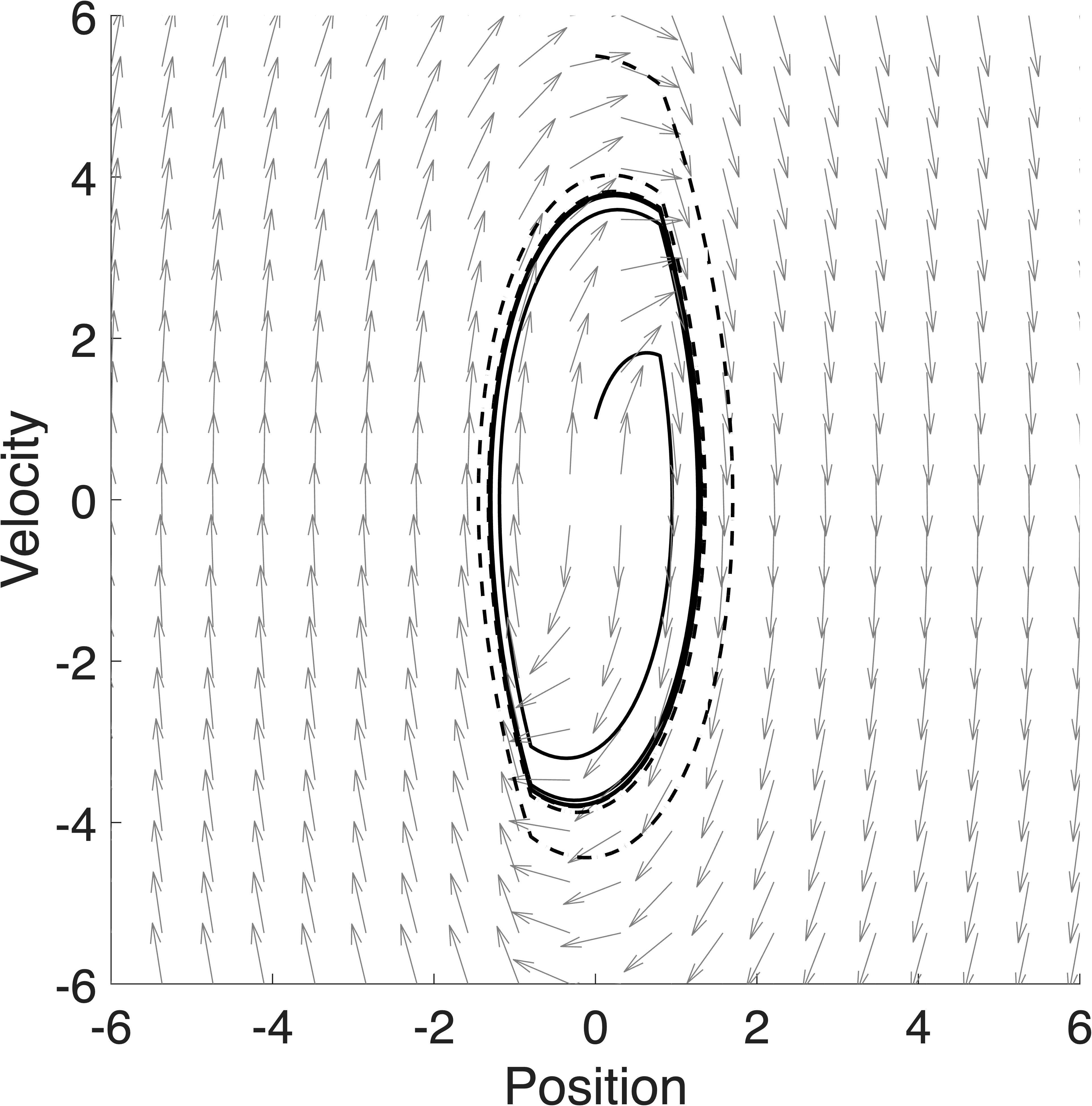}
	\caption{Phase plane of a mass spring damper system with switching references $\pmb{r}_1=-1$ and $\pmb{r}_2=1$ and positions $z_1=-0.8$ and $z_2=0.8$. Trajectories with different initial conditions converge to the same limit cycle.}
	\label{fig:phase_plane}
\end{figure}

Further insights can be obtained by examining a simplified single-degree-of-freedom model, representing a first-order approximation of the vertical dynamics of $\pmb{p}_2$
\begin{equation}
	m \ddot{z} + c \dot{z} + k(z-r) = 0,
\end{equation}
where 
$m>0$ is the mass, $c>0$ is the damping coefficient, and $k>0$ is the spring stiffness. $r$ is the reference, switching between two values, $r \in \{r_1,r_2\}$.
The mechanical energy reads
\begin{equation}
	E = \frac{1}{2} m \dot{z}^2 + \frac{1}{2} k(z-r)^2
\end{equation}
from which
\begin{equation}
	\dot{E} = -c \dot{z} ^2 + W_{r}.
\end{equation}
$W_r$ accounts for the elastic energy injected or removed at switching times (an impulse). 
This can be computed by looking at the left and right limits of the switching time $t$, denoted by $t^-$ and $t^+$,
\begin{equation}
	{E(t^+) \!-\! E(t^-)} = \frac{k{(z(t^+\!)-r(t^+\!))}^2}{2}
	-  \frac{k{(z(t^-\!)-r(t^-\!))}^2}{2}.
\end{equation}
On a period, injection of energy and dissipation must balance.
An example is provided in Fig. \ref{fig:phase_plane}, based on a switching law that captures the fundamental transitions between cutting and raising in the rocking motion:
\begin{align}
	r &= r_2
	\quad\text{if}\quad
	\bigl(\dot{z} \geq 0 \,\text{and}\, z \leq z_2\bigr)
	\;\text{or}\;
	\bigl(\dot{z} \leq 0 \,\text{and}\, z \leq z_1\bigr),
	\nonumber\\
	r &= r_1 
	\quad\text{if}\quad
	\bigl(\dot{z} \geq 0 \,\text{and}\, z \geq z_2\bigr)
	\;\text{or}\;
	\bigl(\dot{z} \leq 0 \,\text{and}\, z \geq z_1\bigr).
	\nonumber
\end{align}

The phase plane of Fig. \ref{fig:phase_plane} shows that solutions converge to a closed curve. 
Intuitively, if the mass moves too fast, the damper dissipates more energy, returning the system toward this closed curve. 
Conversely, if motion is too slow, less energy is dissipated, and the system gains more momentum until it approaches the same closed curve. 

To conclude this section, we provide a practical estimation of the robot's work per cycle. At steady state, this metric quantifies the energetic cost of the cutting process itself (interaction with food and the environment) by isolating it from internal mechanical and virtual damping losses.
\begin{equation}
	E_{\mathcal{C}}
	=
	\int_t^{t+\bar{T}}
	\pmb{F}(\tau)^T \dot{\pmb{x}} (\tau) d \tau,
	\label{eq:work_done_formall}
\end{equation}
where $\bar{T}$ is the cycle period, $\pmb{F}$ is the force at the robot end-effector, and $\pmb{x}$ its position. 
Practically this force is sampled by a force sensor, leading to
the approximated formula
\begin{equation}
	E_{\mathcal{C}_j}
	\ = \!\!\!\!
	\sum_{t_i\in \mathcal{C}_j\setminus\{t_N\}} \pmb{F}^T\bigl(t_i\bigr) \bigl( 
	\pmb{x}(t_{i+1}) - \pmb{x}(t_{i}) \bigr),
	\label{eq:work_done}
\end{equation}
where $\mathcal{C}_j = \{t_1,\dots,t_N\}$ contains the sampling time instants 
of cycle $j$ at which sample measures are taken.

\begin{figure}[htbp]
	\centering
	\includegraphics[width=.75\linewidth]{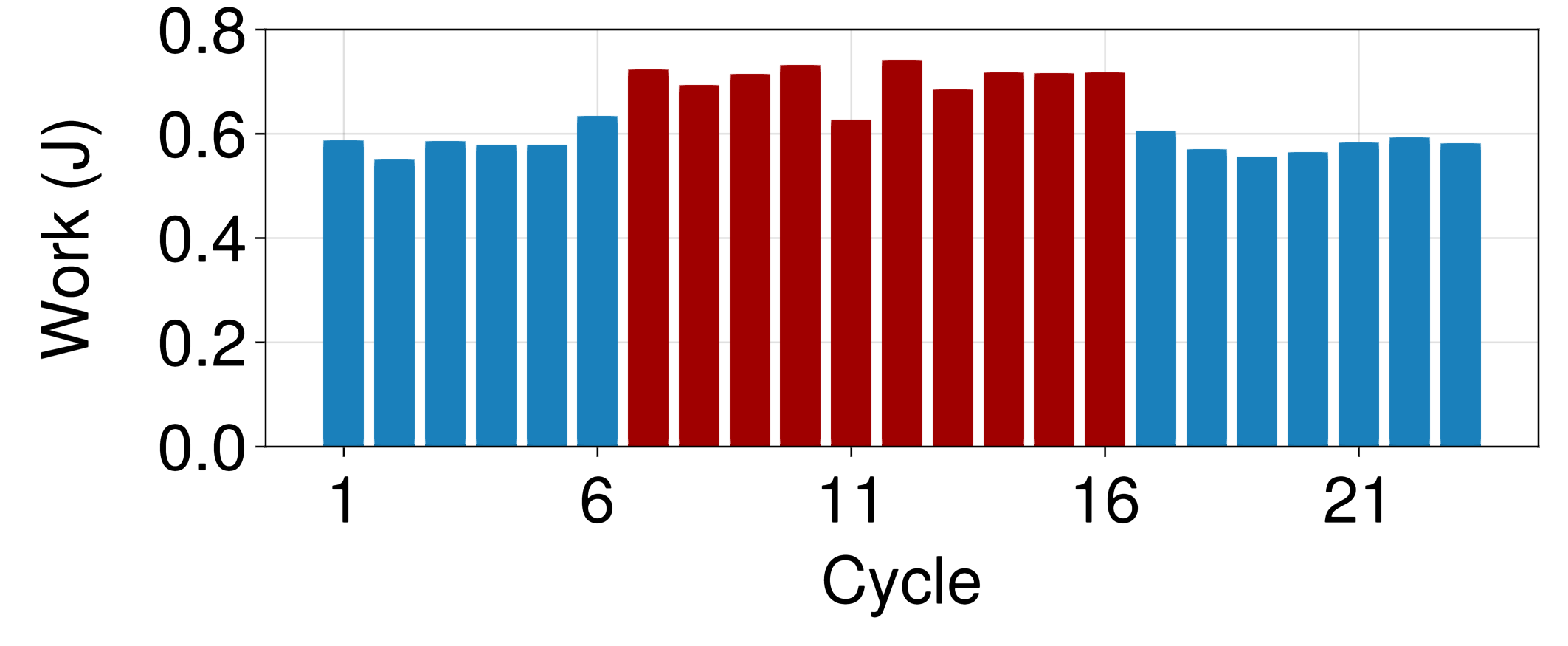}
	\caption{Work done for each cutting cycle. \textcolor{red}{Red}: cutting a carrot, \textcolor{blue}{blue}: no food}
	\label{fig:diff_damping}
\end{figure}

Fig.~\ref{fig:diff_damping} illustrates the work done during a carrot-cutting experiment. Initially, the knife performs a pure rocking motion without food. Between the 7th and 16th cycles, a carrot is introduced and processed, after which it is removed. Fig.~\ref{fig:diff_damping}  provides indirect evidence of the stability of the achieved periodic motion. The work done increases due to the additional dissipation of cutting through the carrot but the system returns to its baseline energy level once the carrot is removed. This behavior demonstrates the virtual model controller's ability to compensate for external perturbations and consistently restore the original energy balance.

\subsection{Stability certificates (numerical)}

Although the energy-based argument above explains why periodic motion can emerge from the balance between switch-induced energy injection and dissipation, it does not provide a full certification of the stability of the periodic motion. This can be achieved through the study of the contractivity of the associated Poincaré map \cite{ParkerChua1989}.

Consider the vector of generalized positions and velocities
\begin{equation}
x =
\begin{bmatrix}
q^\top & \dot q^\top
\end{bmatrix}^\top 
\end{equation}
 where $q \in \mathbb{R}^{m}$ is the aggregate vector of both robot joint positions and VMC generalized coordinates. We take the Poincaré section
 \begin{equation}
\Sigma =
\left\{
x : q_1 = \bar{q}_{1} ,\; \dot q_1 > 0
\right\},
\end{equation}
where $q_{1}$ is the coordinate of the first joint and
$\bar{q}_{1}$ is a specific angle.
 The condition $\dot q_1>0$ identifies the crossing direction.
 Denoting the return map from one crossing of
$\Sigma$ to the next as $P:\Sigma \rightarrow \Sigma$,
a periodic orbit corresponds to a fixed point $x^* = P(x^*)$. 
We can thus obtain a numerical certificate of the asymptotic stability of the periodic attractor by analysing the eigenvalues of the linearized Poincaré map at $x^*$.

The linearization of the Poincaré map at $x^*$ 
  can be derived numerically by considering small perturbations of the form
\begin{equation}
x_i^+ = x^* + \epsilon b_i \quad i \in \{2,…,2m\}
\end{equation}
where each $b_i$ is a tangent vector in the tangent space $T_{x^*}\Sigma$. Specifically, the $i$-th component of $b_i$
is one and all others are zero. The perturbation magnitude is chosen as $\epsilon=0.01$.
Each perturbed state is simulated forward until the next crossing of
$\Sigma$. The $i$-th column of the Jacobian of the Poincaré map is then approximated by 
\begin{equation}
J[:,i] =
\frac{
\left[P(x_i^+)\right]_\perp -
\left[P(x^*)\right]_\perp
}{\epsilon},
\qquad i=1,\ldots,2m-1,
\end{equation}
where $[\cdot]_\perp$ denotes removal of the $q_1$ component. 
$J\in\mathbb{R}^{(2m-1)\times(2m-1)}$ therefore describes the cycle-to-cycle evolution of perturbations tangent to the section $\Sigma$. 
The eigenvalues of $J$ quantify the evolution of perturbations from one cycle to the next. If they are all less than one in magnitude, then perturbations decay over successive cycles, indicating stability of the closed-loop limit cycle. 

We compute this Jacobian for different parameter values to investigate how  parameter variations affect limit cycle stability. As shown in Table \ref{tab:sensitivity}, 
each parameter of the virtual mechanism is varied by $\pm 20\%$ from its nominal value while all other parameters are held fixed. For each case we recompute the Poincaré map Jacobian. Almost all variations lead to stable Jacobians whose eigenvalues have magnitude less than one, indicating asymptotic stability of the corresponding limit cycle. 
The only failed case occurs when the perturbed parameters no longer produce a suitable cutting motion, due to induced fast switching between the two mechanisms.
The stability certificate considers the simulated robot-knife-board system and does not include food fracture dynamics. Robustness to variations in food type and material properties therefore remains an empirical finding, as shown in Section \emph{Experimental result and discussion}.

\section{Experimental setup and implementation}
\label{sec:platform_and_first_experiments}
\subsection{Robot platform and control architecture} \label{implementation}

The virtual model controller outlined in the previous section is implemented through 
the VMRobotControl.jl package \cite{VMRobotControl}, a platform-independent framework designed for virtual model control in robotic systems. 
This package enables the construction and simulation of both the robot and virtual components.
Given the robot's URDF representation,
the package offers several primitives to define the virtual model controller, to compute its dynamics, and to generate the torque commands for each joint.

Our main experimental platform consists of a 7-DoF Franka Research 3 arm, with a maximum payload of 3 kg. 
To collect force data, a ROBOTIQ FT 300-S force-torque sensor is mounted at the robot wrist, providing force and torque measurements at 100 Hz with a noise level of approximately 0.1 N. 
A knife is rigidly connected to the sensor using a 3D-printed attachment as shown in Fig. \ref{fig:experiment_setup}. 
The portability of the virtual model controller allowed us to implement and test it also on the humanoid robot 
Sciurus17, a lightweight and low-cost platform developed by RT Corp. Sciurus17 features 17 degrees of freedom (DoF) distributed across its upper body (with 7 DoF per arm, 2 DoF for the head and 1 DoF for the torso). 
The robot utilizes Dynamixel motors with current control, and the conversion from torque to current is approximated by means of a linear relation based on the performance characteristics of the motor.

\begin{figure}[htbp]
	\begin{center}
		\includegraphics[width=0.62\columnwidth]{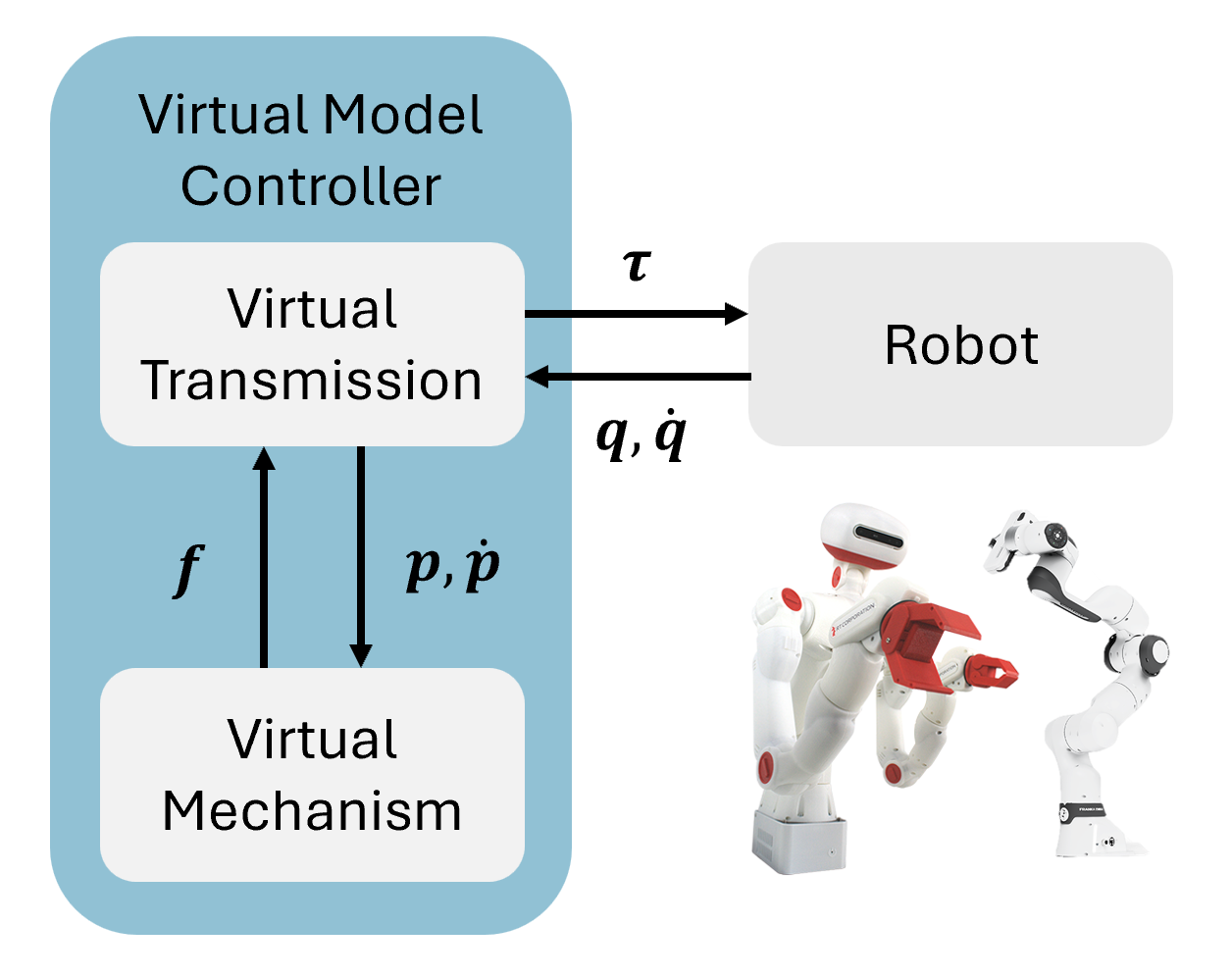}
		\caption{Control architecture implemented with VMRobotControl.jl package.}
		\label{fig:control_architecture}
	\end{center}
\end{figure}

The controller runs on a standard computer.
The overall control architecture is depicted in Fig. \ref{fig:control_architecture}.
Virtual forces $\pmb{f}$ generated by the virtual mechanisms are computed and translated into joint torques $\pmb{\tau}$ using
\eqref{eq:translation}.
The computation of the current state of the virtual mechanisms, such as the elongation of virtual springs and the velocities used by the virtual dampers, is based solely on the robot’s joint position $\pmb{q}$ and velocity $\dot{\pmb{q}}$. 
All force computations and their conversion into joint torques rely on forward kinematics and the associated Jacobians.
This represents a significant departure from traditional motion planning, where execution is typically based on inverse kinematics.

\subsection{The cutting motion in experiments}

\begin{figure*}
	\centering
	\begin{subfigure}[t]{.99\linewidth}
		\includegraphics[width=\columnwidth]{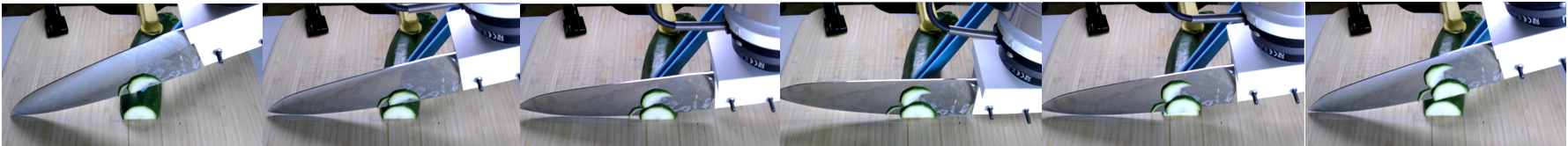}
		\caption{Snapshots from experiment: rocking a courgette with a knife.}
		\label{fig:exp_1_1_snapshots}
	\end{subfigure}
	\begin{subfigure}[t]{.55\linewidth}
		\includegraphics[width=\columnwidth]{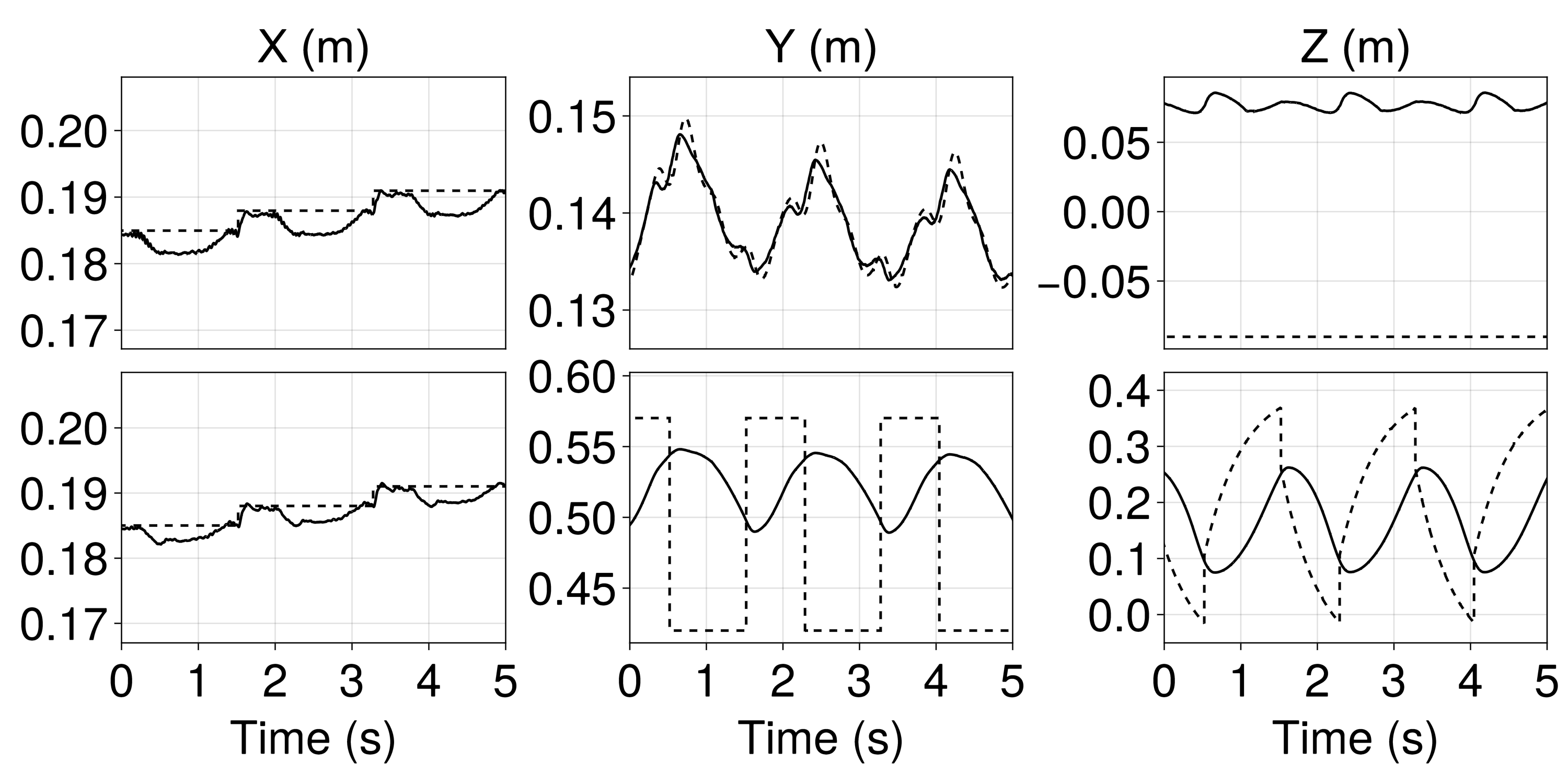}
		\caption{Top: Solid: $\pmb{p}_1$, Dashed: $\pmb{p}_a$. Bottom: Solid:  $\pmb{p}_2$, Dashed: $\pmb{p}_b$.}
		\label{fig:rocking_1_1_tip_back}
	\end{subfigure}
	\begin{subfigure}[t]{.49\linewidth}
		\includegraphics[width=\columnwidth]{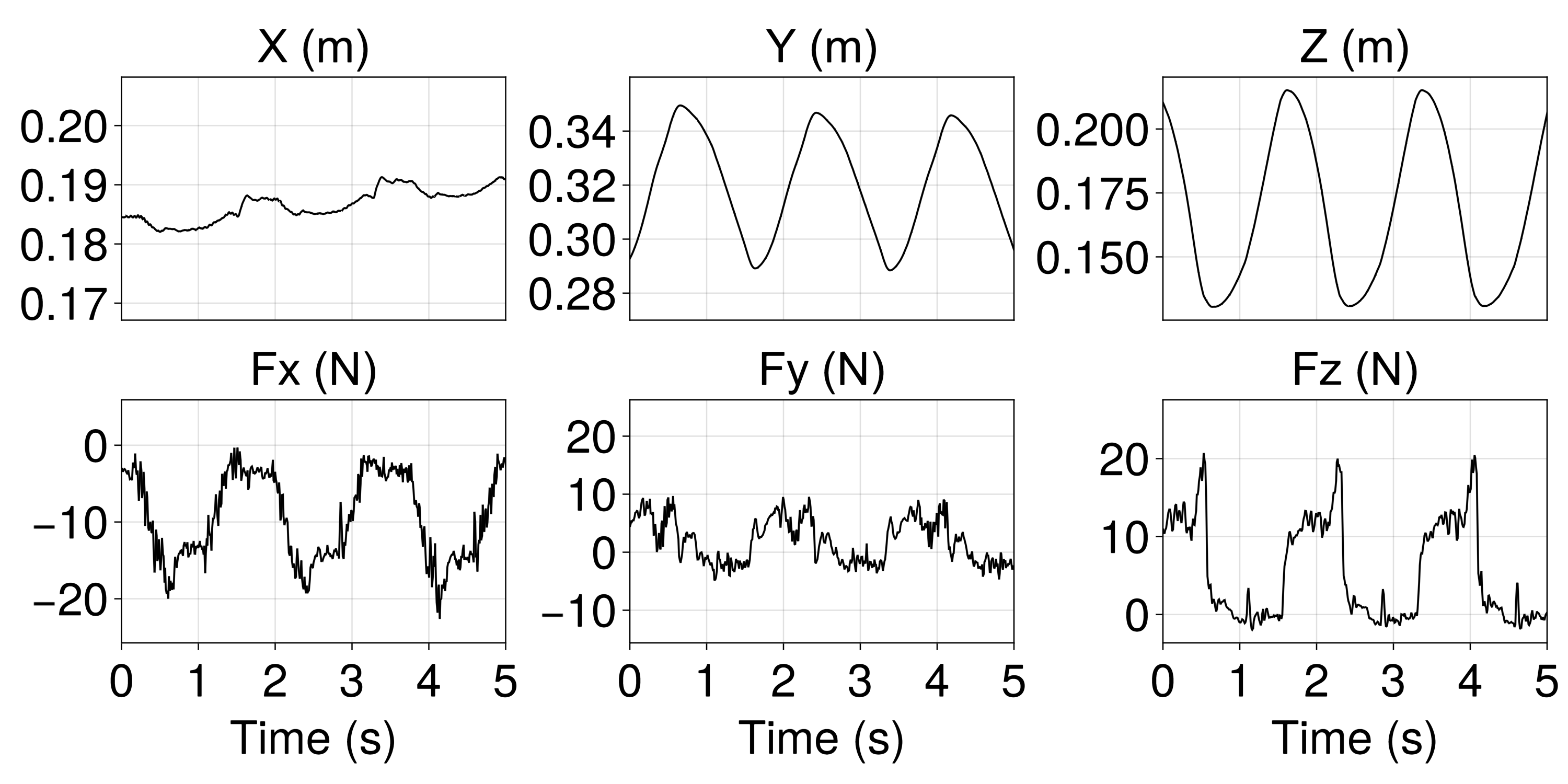}
		\caption{End effector position and force reading.}
		\label{fig:rocking_1_1_EE}
	\end{subfigure}
	\begin{subfigure}[t]{.49\linewidth}
		\includegraphics[width=\columnwidth]{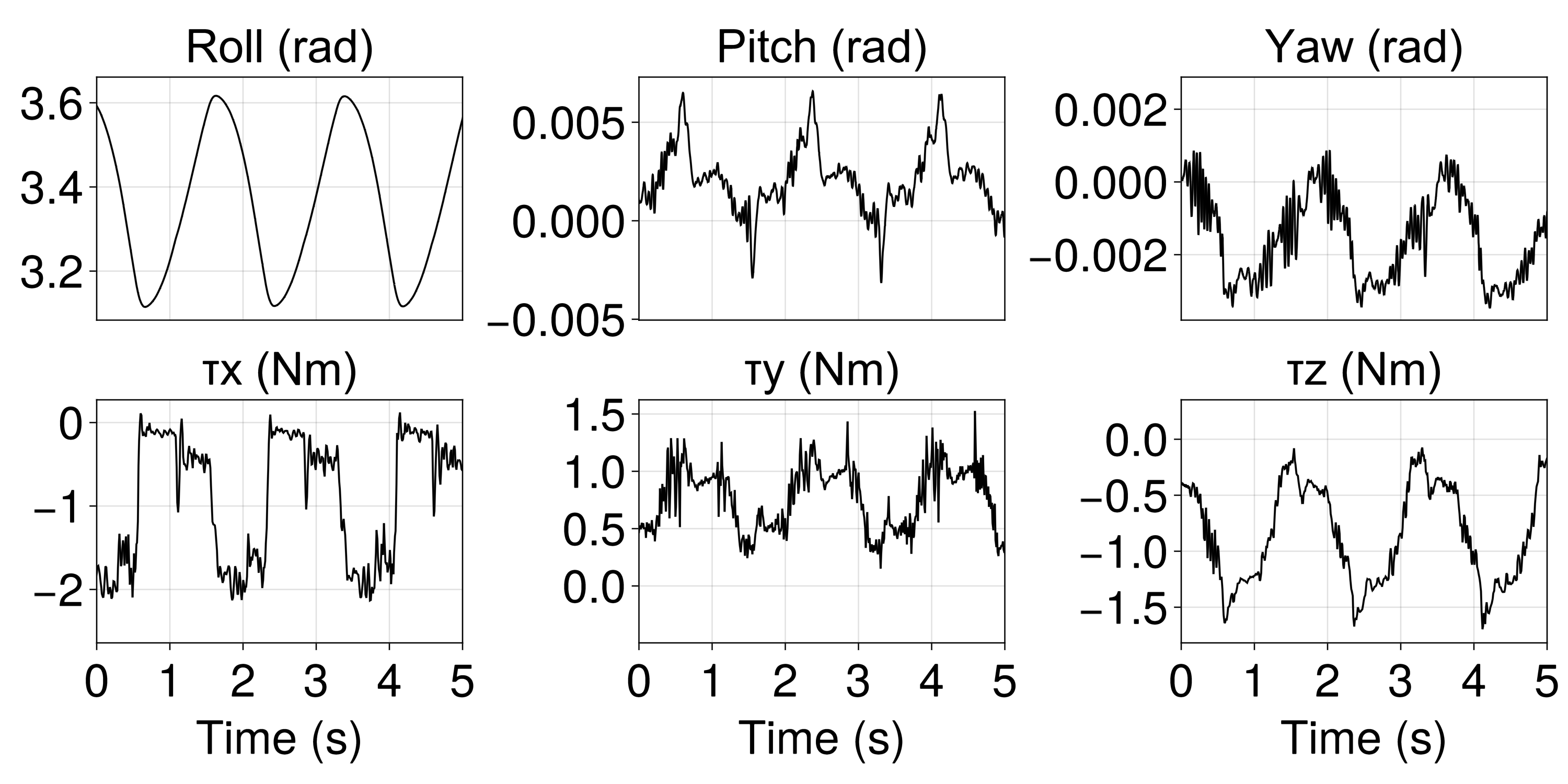}
		\caption{End effector orientation in Euler angles and torque reading.}
		\label{fig:rocking_1_1_EE_euler}
	\end{subfigure}
	\caption{Rocking motion as a result of the interaction between the virtual mechanism and environment.}
	\label{fig:exp_1_1_data}
\end{figure*}

The virtual mechanisms constrain the motion of the blade by enforcing compliance at contact points and `soft' constraints. Switching of the reference at specific points induces a cyclical motion of the knife, which is determined by the interaction among robot dynamics, compliant virtual mechanism constraints, and contacts with the environment. A first illustration is provided in Fig. \ref{fig:exp_1_1_data},
whose control parameters are specified in Table \ref{tab:control_parameters} (a).
Using courgette as the cutting target, the snapshots in Fig. \ref{fig:exp_1_1_snapshots} illustrate the continuous contact between the knife and the cutting board while the knife rolls back and forth. 
The successful separation of the courgette is also seen in the snapshots. Since the motion is not pre-programmed,
variations in the knife geometry or cutting board features will yield different motions.

Fig. \ref{fig:rocking_1_1_tip_back} shows the position of two springs responsible for the rocking motion.
The sliding of the virtual mass ensures that $\pmb{p}_a$ stays close to $\pmb{p}_1$ and no additional force is applied in the rocking direction.
The displacement between $\pmb{p}_1$ and $\pmb{p}_a$ along the $z$ axis pulls the knife towards the cutting board and ensures contact.
The knife moves to the next slice position when raised. 
Fig. \ref{fig:rocking_1_1_EE} and \ref{fig:rocking_1_1_EE_euler} show the end-effector's position and force, orientation and torque. The resulting end-effector's motion is a fairly complicated trajectory, difficult to design manually.
One would need to know the geometry of the knife to decide where the end-effector should be while rolling the knife back and forth.

Force and torque are measured at the robot's wrist and transformed to the world coordinate.
$F_y$ and $F_z$ correspond to the forces applied to the robot wrist to slice and cut.
$F_x$, $\tau_y$ and $\tau_z$ act to stabilize the knife on the slicing plane.
The pitch and yaw remain small, which shows that the stabilization of the slicing plane is effective. 
The roll and torque $\tau_x$ account for the rocking motion.
During the lifting of the knife, a higher amount of torque $\tau_x$ is applied when the contact point is shifted further to the tip.

\section{Experimental result and discussion} 
\label{experiments}

\subsection{Forces and velocities}
\label{cutting_with_vm}
\begin{figure*}
	\centering
	\begin{minipage}[h]{0.6\textwidth}
		
		\begin{subfigure}[t]{\linewidth}
			\includegraphics[width=\linewidth]{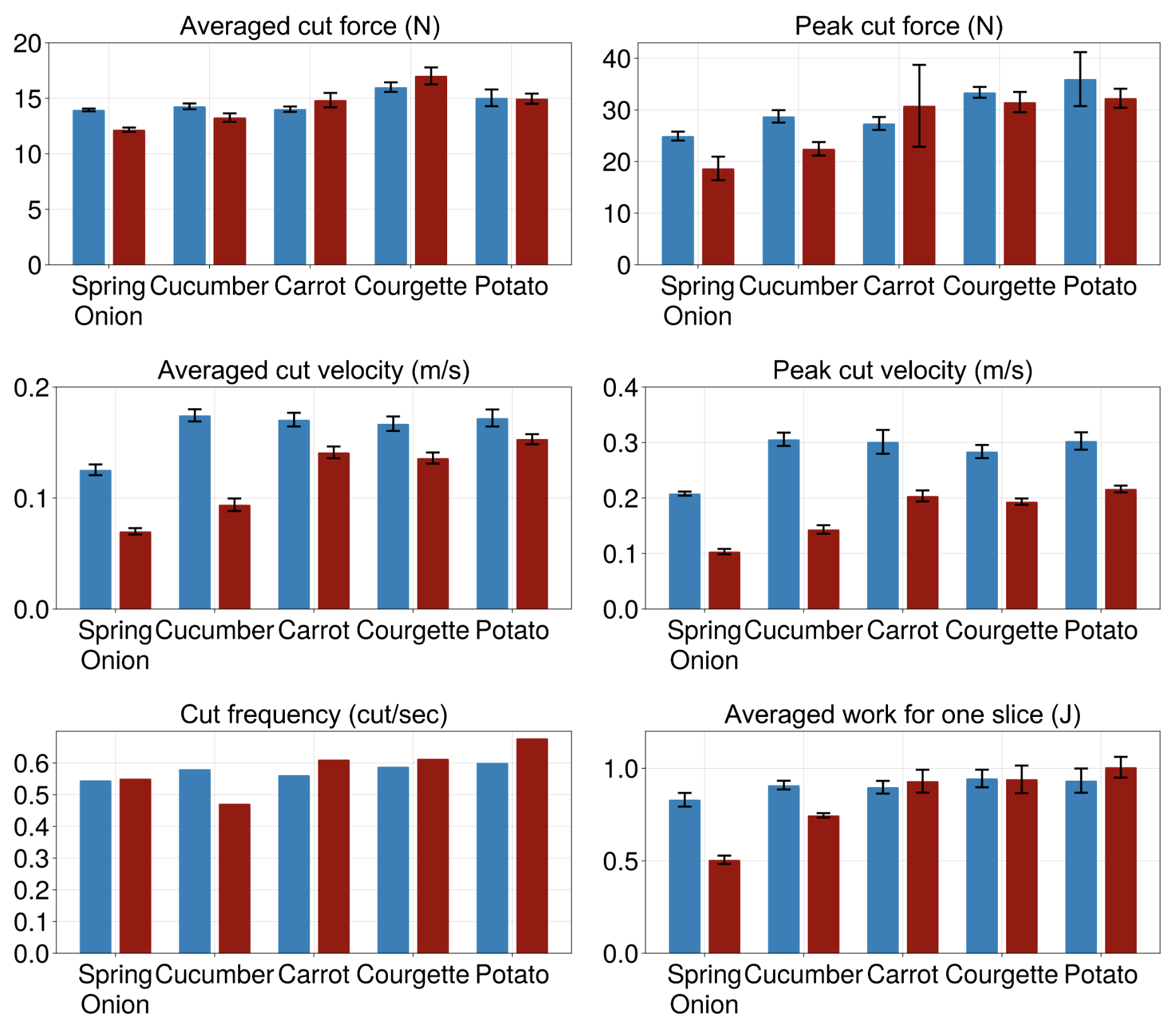}
		\end{subfigure}
		
	\end{minipage}
	\hspace*{\fill}
	\begin{minipage}[h]{0.38\textwidth}
		\begin{subfigure}[t]{\textwidth}
			\includegraphics[width=\linewidth]{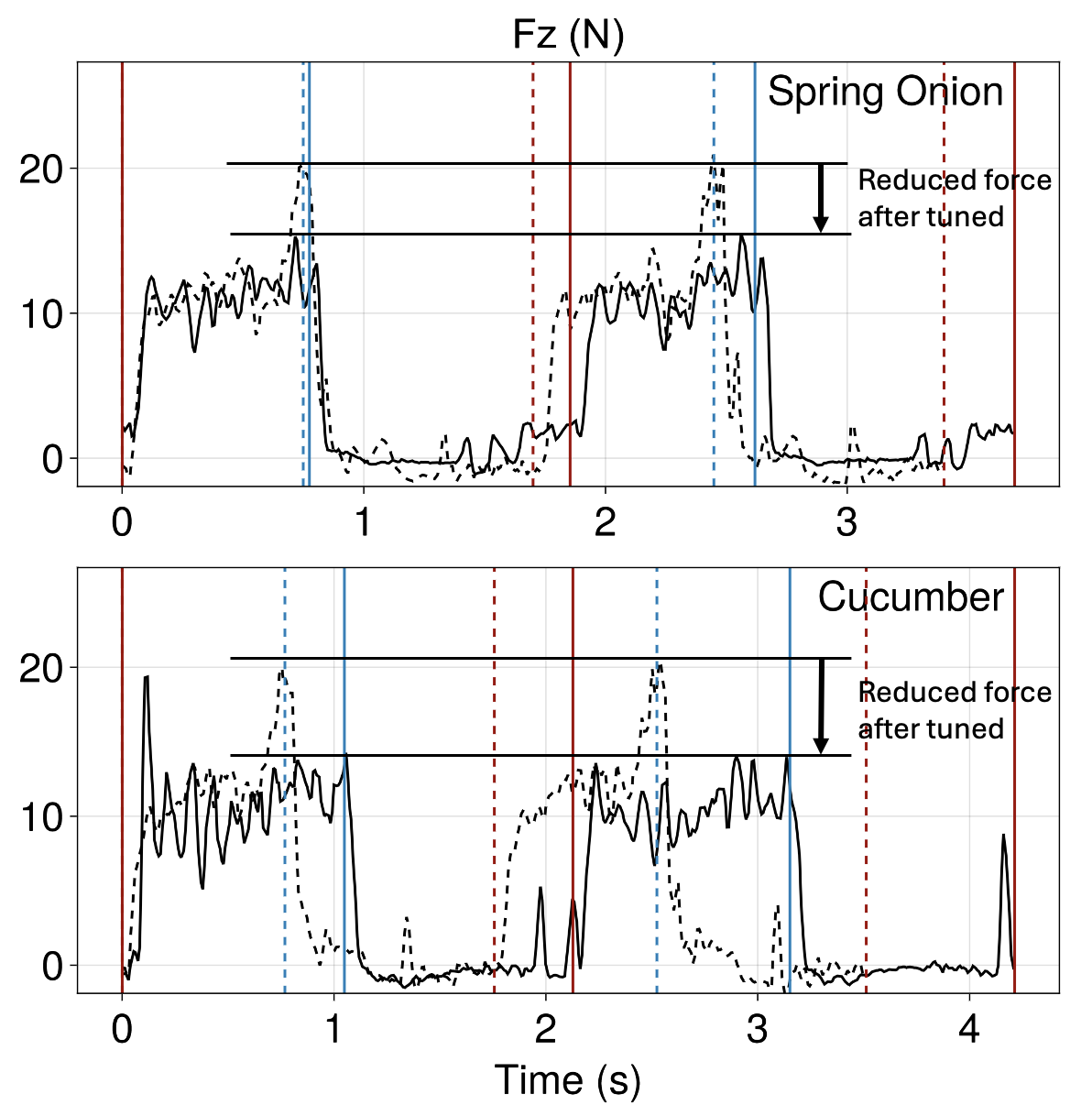}
			\caption{Black: Force reading in the z axis. Red: Switch from raising to cutting. Blue: Switch from cutting to raising. Dash: Same control parameters for all food. Continuous: Tuned control parameters for each food so that the knife cuts through without applying excessive force.}
			\label{fig:exp_1_diff_food_forcez}
		\end{subfigure}
	\end{minipage}
	\caption{Rocking mechanism applied to five different food. Left two columns: \textcolor{blue}{blue} - untuned, \textcolor{red}{red} - tuned for minimal force.}
	\label{fig:exp_1_diff_food_rocking_cutting_properties}
\end{figure*}

To evaluate the controller's versatility, we conducted cutting experiments on a variety of vegetables under two conditions: (i) a \emph{uniform parameter set} applied to all food types, and (ii) \emph{food-tuned parameters} manually optimized to minimize contact force.
Reducing the cutting force helps to achieve precise cuts and prolongs the lifespan of both the knife and the cutting board.
The tested objects span distinct plant-tissue structures, including thin fibrous tissue (spring onion), soft high-moisture tissue (cucumber and courgette), crisp root vegetable tissue (carrot), and tuber tissue (potato), with their distinct properties reported in \cite{VINCENT2004695,Zdunek2008TurgorAT}. These differences are expected to influence deformation, fracture, and energy dissipation during cutting.
Larger items, such as cucumber, courgette, and potato, were cut in half before the experiments.
This reflects the intended use of the rock-chop technique, which is most suitable for smaller objects that allow the curved blade to roll while maintaining contact with the cutting board.
\textcolor{black}{
All vegetables were securely held using a clamp and tongs to prevent motion during cutting. The tongs restrained the food as the cut progressed, serving a role analogous to the supporting hand during manual cutting, without contacting the blade or the region being cut. This resulted in a downward force that helped to hold the food in place, possibly facilitating complete separation, thereby contributing to the 100\% success rate reported in Table \ref{tab:vegetable_success} and Fig. \ref{fig:exp_2_height_result}.}
Results are summarized in Fig. \ref{fig:exp_1_diff_food_rocking_cutting_properties}.
A cut is considered successful when the vegetable is completely separated. Across all five vegetable types, all attempted cuts were successful under both the uniform and food-tuned parameter settings. The corresponding success counts and rates are reported in Table \ref{tab:vegetable_success}.
The rocking motion was initially validated through simulation but parameter tuning was performed on the physical robot to account for the complexities of modeling contact forces during cutting. Although we explore the influence of these parameters on performance, a systematic optimization study remains beyond the scope of this work. We refer the reader to \cite{larby_optimal_2025} for an optimization framework tailored to virtual model control.

\begin{table}[t]
\centering
\footnotesize
\setlength{\tabcolsep}{2.5pt}
\renewcommand{\arraystretch}{0.9}
\begin{tabular}{lcc}
    \toprule
    Vegetable & Uniform parameters & Tuned parameters \\
    \midrule
    Spring onion & $10/10$ (100\%) & $10/10$ (100\%) \\
    Cucumber     & $10/10$ (100\%) & $10/10$ (100\%) \\
    Courgette    & $10/10$ (100\%) & $10/10$ (100\%) \\
    Carrot       & $10/10$ (100\%) & $10/10$ (100\%) \\
    Potato       & $10/10$ (100\%) & $10/10$ (100\%) \\
    \bottomrule
\end{tabular}
\vspace{2mm}
\caption{Cutting success across different vegetables under uniform and tuned parameter settings. Each entry reports the number of successful cuts over the total number of attempted cuts, followed by the corresponding success rate. A cut is considered successful when the vegetable is completely separated.}
\vspace{-2mm}
\label{tab:vegetable_success}
\end{table}

When using a single set of control parameters for all foods, the force profiles in the $z$-axis (Fig. \ref{fig:exp_1_diff_food_forcez}) were similar across all food types despite significant differences in the material properties of different food.
This is because the control parameters were chosen to apply a sufficiently high force so that the intrinsic properties of the food had minimal impact on the resulting behavior.
Peak force is typically observed near the transition from cut to raise and varies with the energy absorption characteristics of each vegetable.
Higher levels of `residual' energy after cut typically results in a higher contact forces before the knife is retracted. 
With parameters tuned specifically for each food type, the force required for cutting spring onion and cucumber is noticeably reduced, while high force levels are needed for other vegetables.
In terms of tuning strategy, stiffness of springs was uniformly reduced (increased) to reduce (increase) contact forces. The cut transition parameters ($m_b, k_b, c_b$) were adjusted accordingly to preserve cutting frequency. 
For smaller items, the knife’s lifting height was reduced by increasing the switching tolerance $\delta_2$, maintaining comparable cutting frequencies even with lower cutting velocities.
For larger items, the reference position for raising the knife, $\pmb{r}_{2,2}$, was set higher to accelerate the upward motion, with $\delta_2$ adjusted accordingly.
Compared to the uniform parameter, cut frequency remains similar or becomes higher while velocity during cutting is reduced for carrot, courgette and potato. 

These results do not provide an exhaustive characterization,  which is beyond the scope of the current paper, as this would require a proper optimization framework and extensive validation tests for a proper statistical analysis, further complicated by the extreme variability in mechanical properties of vegetables, \cite{2003toughnessOfFood}. However, they illustrate the tunability of the virtual model controller, by showing how
different parameter configurations can achieve successful cutting while producing distinct motion characteristics, allowing the control designer to shape the desired behavior through physical parameter selection rather than trajectory specification.
Fig. \ref{fig:slices_of_veges} shows the sliced vegetables at the end of experiments.  By visual inspection, both (i) uniform and (ii) food-specific parameter selection 
produce no noticeable difference in slice quality or accuracy.

\begin{figure*}[t]
	\centering
	\includegraphics[width=.9\linewidth]{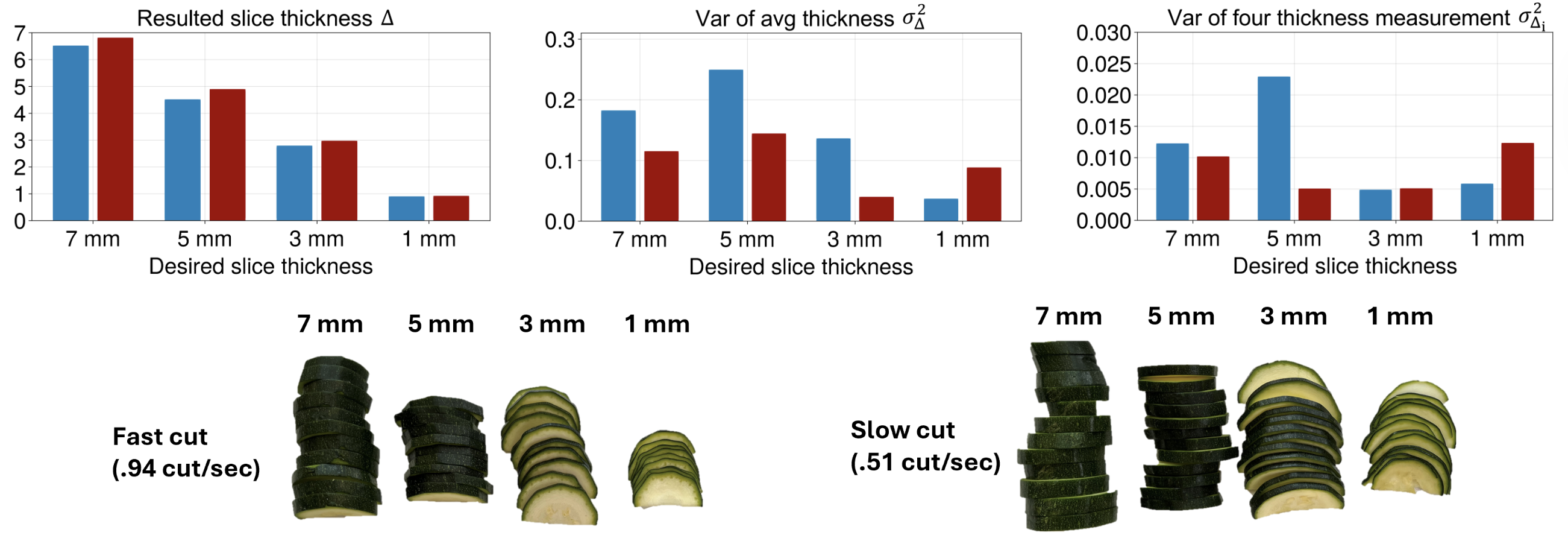}
	\vspace{-2mm}
	\caption{Cutting different slices. Top row: \textcolor{blue}{blue} - 0.94 cut/sec, \textcolor{red}{red} - 0.51 cut/sec. Bottom row: fast and slow frequencies.}
	\label{fig:exp_3_metrics}
\end{figure*}

\begin{figure*}[ht]
	\centering
	\begin{subfigure}[t]{\linewidth}
		\centering
		\includegraphics[width=\columnwidth]{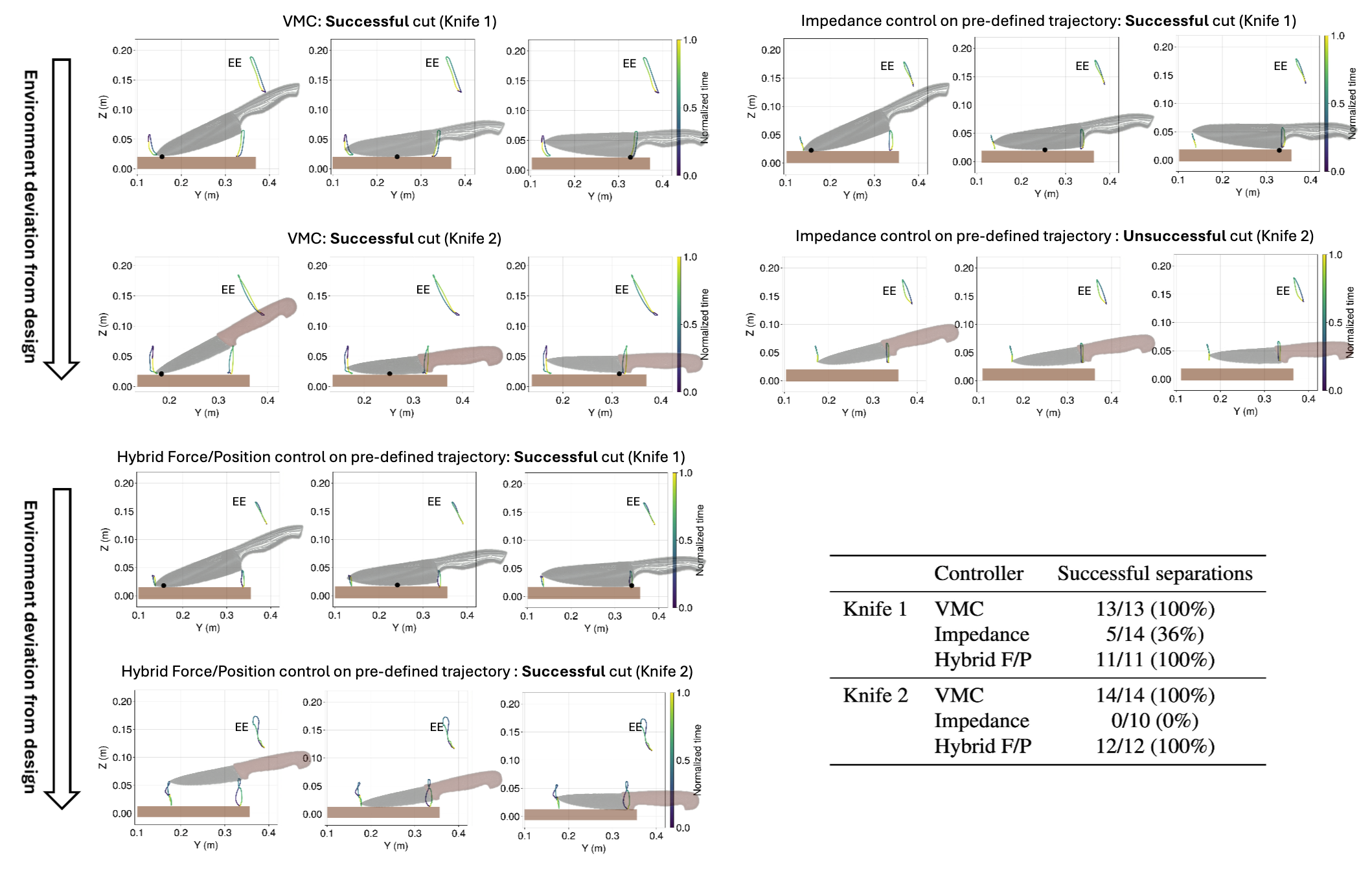}
        
		\caption{
        Rocking with the proposed VMC succeeds for knives with different sizes and blade profiles. In contrast, impedance control following a pre-defined trajectory fails under variations in knife geometry. The reference trajectory is the end-effector position and orientation recorded during a successful cut with Knife 1 under VMC. Trajectories of the knife tip, heel and the end-effector are shown in each figure.
        }
		\label{fig:knife_traj}
	\end{subfigure}
	
	\vspace*{1mm}
	
	\begin{minipage}[h]{1\textwidth}
		\centering
		\begin{subfigure}[t]{1\linewidth}
			\centering
			\includegraphics[width=0.9\linewidth]{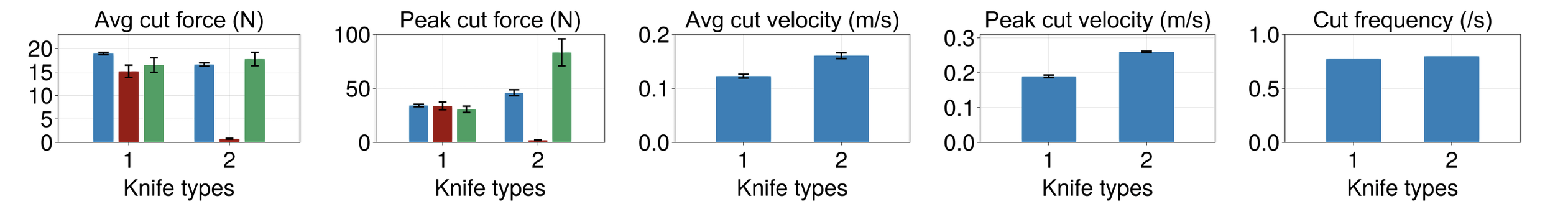}
			\caption{Forces, velocities, and cut frequencies for different knives. \textcolor{blue}{Blue}: cutting with the proposed VMC; \textcolor{red}{red}: impedance control on predefined trajectory; \textcolor{green}{green}: hybrid force/position control on predefined trajectory.}
            \label{fig:exp_2_knife_metrics}
		\end{subfigure}
	\end{minipage}
	\hspace*{\fill}
	\caption{Rocking motion with different knives \textbf{using the same virtual mechanism and control parameters}.}
	\label{fig:exp_2_knife_result}
\end{figure*}

\subsection{Accuracy}

To evaluate the precision of our cutting mechanism, we cut courgette samples using various target thicknesses and cutting speeds. Two average cut frequencies were implemented for each thickness setting, approximately 0.51 and 0.94 cuts per second. Fig. \ref{fig:exp_3_metrics} shows representative slices obtained under these conditions.

Cutting accuracy is assessed by measuring each cut in four distinct locations. Measurements are taken with care, avoiding deformation of the slice as much as possible.
We use three metrics to measure (i) if the desired thickness is achieved, (ii) if the accuracy is consistently achieved with continuous cutting, and (iii) if each slice is evenly cut.

For (i),  we compare the average of all measurements $\Delta$ with the desired one. We compute
\begin{equation}
	\Delta = \frac{1}{N}\sum_{i=1}^{N} \Delta_{i}
	\qquad\quad 
	\Delta_i = \frac{1}{4}\sum_{j=1}^{4} \Delta_{i,j}
\end{equation}
where $N$ is the total number of slices, $\Delta_{i}$ is the average thickness of slice $i$ based on four thickness measurements 
$\Delta_{i,j}$.
For (ii), 
we take the variance of the average thickness slices. Namely,
\begin{equation}
	\sigma^2_{\Delta} = \frac{1}{N} \sum_{i=1}^{N} \left(\Delta_i - \Delta\right)^2.
\end{equation}
Higher variance indicates lower consistency. Finally, 
for (iii), we take the variance of the four thickness measurements within each slice and compute its average across all slices:
\begin{equation}
	{\overline\sigma^2_\Delta} = \frac{1}{N} \sum_{i=1}^{N} \sigma^2_{\Delta_i}
	\qquad
	\sigma^2_{\Delta_i} = \frac{1}{4} \sum_{j=1}^{4} \left( \Delta_{i,j} - \Delta_i \right)^2.
\end{equation}
In this case, a higher variance indicates lower evenness.

Fig. \ref{fig:exp_3_metrics} shows that average thickness is close to the desired target. In particular, the slices are consistently slightly thinner than expected. 
Several factors may contribute to this discrepancy: the knife's movement matches the desired thickness without compensating for its own width, alongside the plastic deformation of the vegetable during cutting and moisture loss in the slices.
Despite these effects, achieving nearly one slice per second with such precision is comparable to or even exceeds human performance.

Faster cutting generally degrades performance in terms of average thickness, consistency, and evenness, as illustrated in Fig. \ref{fig:exp_3_metrics}, except in the case of 1-mm slices. 
The decreased performance at higher speeds is mainly due to the limited ability of the knife to accurately reach the target slicing position in a short time frame, resulting in fluctuations and positional adjustments during the cut. 
This effect is minimized when cutting very thin slices. Additionally, 1-mm slices experience greater plastic deformation at slower cut speeds, potentially hindering the initiation and propagation of a clean cut.

\begin{figure}[h]
	\centering
	\begin{subfigure}[h]{.95\linewidth}
		\includegraphics[width=\columnwidth]{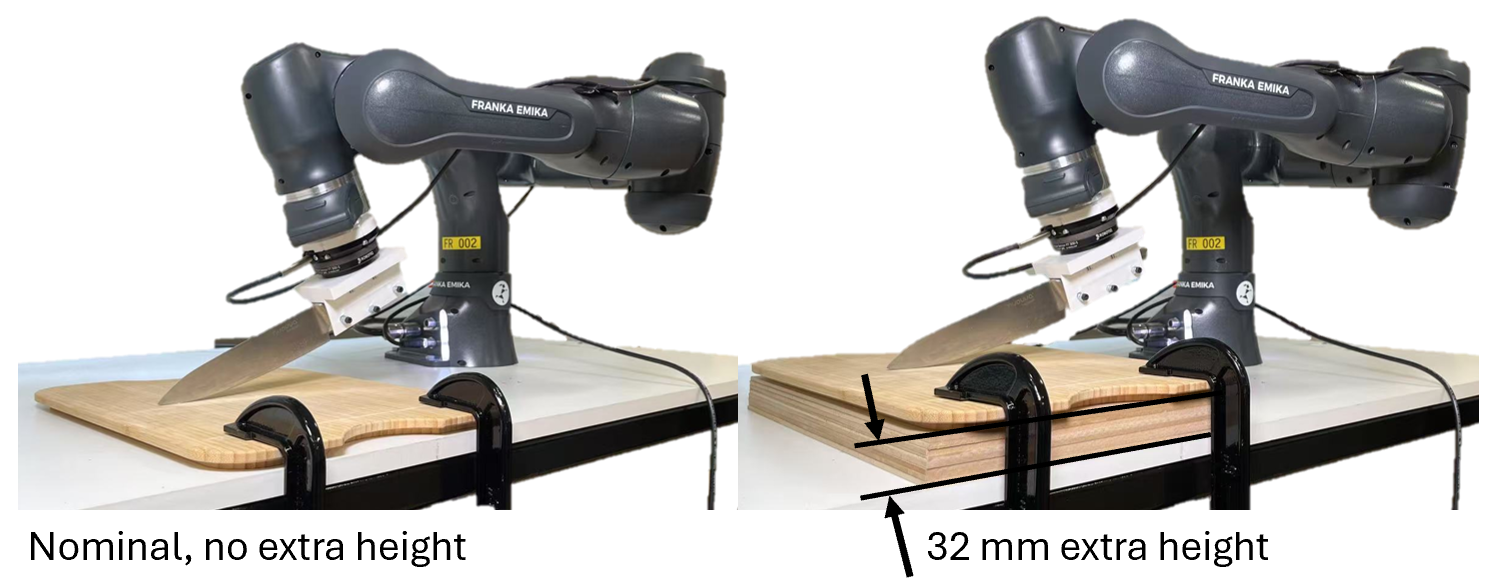}
		\caption{Experiment setup with different cutting board heights.}
		\label{fig:rocking_2_height_snapshots}
	\end{subfigure}
	
	\vspace*{2mm}
	
	\begin{subfigure}[t]{1\linewidth}
		\centering
		\includegraphics[width=\columnwidth]{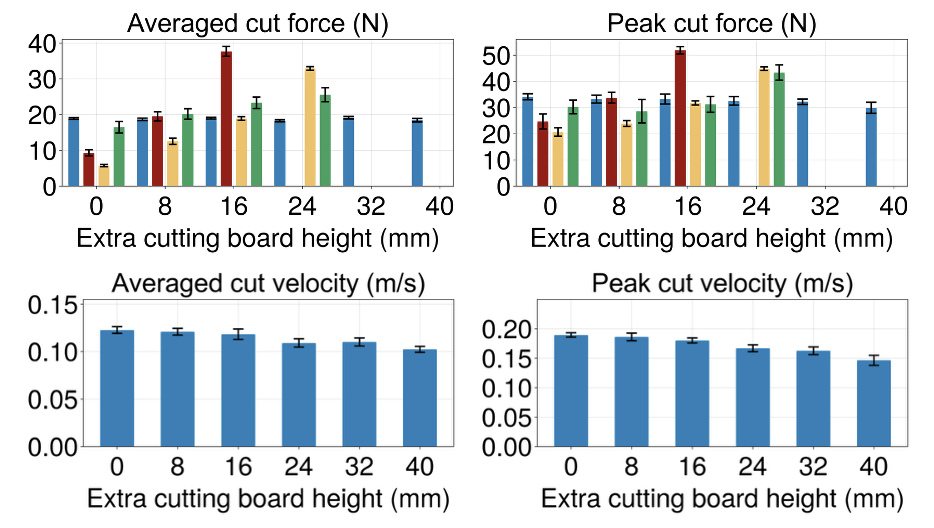}

        {
        \scriptsize
        \setlength{\tabcolsep}{2.5pt}
        \renewcommand{\arraystretch}{0.9}
        \begin{tabular}{lcccc}
            \toprule
            \makecell{Extra height\\{[mm]}} & VMC & \makecell{Impedance\\(high)} & \makecell{Impedance\\(low)} & Hybrid F/P \\
            \midrule
            0  & 13/13 (100\%) & 5/14 (36\%)  & 0/11 (0\%)   & 11/11 (100\%) \\
            8  & 13/13 (100\%) & 14/14 (100\%) & 14/14 (100\%) & 11/11 (100\%) \\
            16 & 12/12 (100\%) & 13/13 (100\%) & 14/14 (100\%) & 7/7 (100\%) \\
            24 & 13/13 (100\%) & 0/1 (0\%)   & 13/13 (100\%) & 10/10 (100\%) \\
            32 & 13/13 (100\%) & 0/1 (0\%)   & 0/1 (0\%)   & 0/1 (0\%) \\
            40 & 12/12 (100\%) & 0/1 (0\%)   & 0/1 (0\%)   & 0/1 (0\%) \\
            \bottomrule
        \end{tabular}
        }
        
		\caption{
        Forces and velocities for different cutting-board heights. The table reports the number of successful separations over the total number of cuts. Impedance control fails to complete the motion at a +24 mm height with high gains and at +32 mm with low gains due to excessive contact forces on the end-effector, while the proposed VMC continues to operate. \textcolor{blue}{Blue}: cutting with the proposed VMC; \textcolor{red}{red}: impedance control (high gain); \textcolor{yellow}{yellow}: impedance control (low gain); \textcolor{green}{green}: hybrid force/position control on predefined trajectory.}
	\end{subfigure}
	\caption{Rocking with different cutting boards \textbf{using the same virtual mechanism and control parameters}.}
	\label{fig:exp_2_height_result}
\end{figure}

\subsection{Robustness}

Our virtual model controller demonstrates inherent robustness. As shown in previous section, the same set of control parameters achieved successful cut of a variety of food items, from delicate spring onions to rigid potatoes. 
In addition to these results, we tested the controller using two different knives.
Knife 1 features an optimal blade curvature for rocking and has been used in all experiments so far.
Knife 2 has a different blade geometry, which is still suitable for rocking.
Results are shown in Fig. \ref{fig:exp_2_knife_result}. To assess robustness, we consider a uniform set of parameters for both knives.
Although both knives exhibited rocking motion, the differences in blade geometry affected energy dissipation and resulted in different rocking features.
Knife 2 has a smaller blade than Knife 1, resulting in a reduced energy dissipation through contact with the cutting board. This led to a higher residual energy after cutting, which turned into speed, that is, faster cuts.
The increase in force peaks for Knife 2 can be explained in a similar way: Knife 2 moves faster than Knife 1, leading to stronger forces around the transition from the cutting phase to the raising phase.

We compared our approach against a standard impedance controller and a hybrid force/position controller, both based on a pre-defined nominal trajectory. 
Deriving a suitable rocking trajectory is itself non-trivial, as it requires precise knowledge of the knife geometry and cutting board height. 
We therefore constructed the nominal trajectory from the end-effector trajectory of a successful cutting motion with Knife 1 generated by the virtual model controller.
For the impedance baseline, a Cartesian impedance controller with high gains (Table~\ref{tab:control_parameters} (d)) was used to track the recorded pose trajectory. 
The gains were tuned manually to achieve the best tracking accuracy without inducing high-frequency oscillations.
The hybrid force/position baseline is designed to represent an extension of control framework in \cite{2023YanBin}, adapted to our experimental setting. 
In \cite{2023YanBin}, cutting is decomposed into pressing, touching, and slicing stages: the knife is first driven toward the cutting board using Cartesian position control, impact with the board is handled using hybrid position--impedance control, and the subsequent slicing motion is controlled using hybrid position--force control. 
Since our task requires repeated cut cycles rather than a single cut, we use the same hybrid force/position control during the cutting stage, while adding a separate knife-raising phase for repeated operation.
During the cutting stage, the baseline follows the nominal rock chop trajectory, but replaces vertical position tracking with force regulation. 
Specifically, the closed-loop Cartesian force command along the $z$-axis reads
\begin{equation}
    F_z^{\mathrm{cmd}}(k) = F^{\mathrm{des}}_z +
    k_{I} \sum_{j=k_{\mathrm{cut}}}^{k}
    \left(F^{\mathrm{des}}_z - F_z(j)\right)\Delta t ,
\end{equation}
where $F_z$ is the approximated contact force obtained from measured force at the robot's wrist after transformation to the world frame, $F_z^{\mathrm{des}}$ is the desired force, $k_{I}$ is the integral force gain, and $k_{\mathrm{cut}}$ denotes the first sample time of the cutting stage. 
We set $F_z^{\mathrm{des}}=20\,\mathrm{N}$, based on the interaction force observed in successful cutting trials generated by our virtual mechanism controller.
For the knife-raising motion, the baseline switches back to Cartesian pose tracking of the recorded trajectory using impedance control. 
The $z$-axis impedance gain in the hybrid force/position baseline was reduced relative to the pure impedance baseline to smooth the transition between impedance-based trajectory tracking and cutting-stage force regulation.

For both comparisons, we define the cutting phase as the time interval during which the end-effector $z$-position is decreasing. Cut forces are then computed using \eqref{eq:force_computation}.

The impedance baseline could reproduce the recorded rocking motion and cutting was generally successful, although noticeable tracking discrepancies remained. Notably, a thin film of onion at the base was left unseparated in most of the cuts, indicating insufficient normal cutting force. With Knife 2, the geometric mismatch of the knife shape caused the blade to lose contact with the board, leading to poor cuts without complete separation.
The hybrid force/position baseline adjusts the vertical motion online based on the measured force error, thereby maintaining contact with the cutting board and providing sufficient interaction force for material separation. 
This baseline successfully cuts through the spring onion. 
However, because it still replays the recorded pose trajectory during the knife-raising stage, the trajectory lifts Knife 2 excessively and causes loss of contact with the cutting board. 
When the controller switches back to the cutting stage, the knife re-establishes contact abruptly, resulting in the large impact-force peak observed for Knife 2 in Fig. \ref{fig:exp_2_knife_metrics}.
Thus, although both hybrid force/position and VMC controllers achieve successful separation under the tested knife-geometry variation, VMC generates the rocking motion without requiring a predefined trajectory and avoids the high peak force due to trajectory mismatch.

We also demonstrate the ability of the virtual model controller to handle different heights of the cutting board. Fig. \ref{fig:rocking_2_height_snapshots} illustrates the experimental setup, where the nominal cutting board height was modified by adding MDF layers underneath. 
Fig. \ref{fig:exp_2_height_result} summarizes the force, velocity, and success rate for the different cutting-board heights. The table reports the number of successful separations over the total number of attempted cuts. 
The number of cuts varies across experimental conditions because the available number of consecutive slices depended on the usable length of each vegetable sample.
Without changing any control parameters, the virtual model controller maintained a stable and successful rocking motion across all heights.
As the board height increased, the peak cutting force and peak and average cut velocities gradually decreased, which can be explained by the reduced distance available for the knife to accelerate before switching and by the saturating profiles of the nonlinear springs.

Also in this scenario, we compared our approach against the same two  baselines. For impedance control, both high-gain and low-gain settings were tested (Table \ref{tab:control_parameters} (d)). For hybrid force/position control, the recorded trajectory was used with the force control in the $z$ axis. The impedance controller tracked the pre-defined nominal trajectory constructed as specified above. For the cutting board at the nominal height, the low-gain impedance controller failed to cut onions completely, with the cut approximately reaching halfway through due to insufficient applied force. When the cutting board was raised by 8 mm, the same impedance controller was then able to cut through the spring onions, as the higher board effectively increased penetration for a given trajectory. 
However, further increases in board height led to larger contact forces, triggering protective stops due to excessive forces on the arm. Specifically,  the high-gain controller failed at 24 mm height increment, while the low-gain controller tolerated higher displacements but ultimately failed at 32 mm increment.
The hybrid force/position baseline tolerated larger cutting-board height variations than the impedance baselines, because the $z$-axis force is regulated during the cutting stage. 
Since the baseline replays the recorded Cartesian pose during knife raising, a higher board creates increasing inconsistency with the actual board location. 
While the average force during cutting increased with board height, this trend is consistent with the finite settling time of the integral force controller. 
The tighter force regulation reported in \cite{2023YanBin} is consistent with their slower cutting regime, which gives the integral force controller more time to settle.
For the \(32\,\mathrm{mm}\) and \(40\,\mathrm{mm}\) height increments, this mismatch produced excessive contact forces and triggered the robot's protective stop. 
For the \(16\,\mathrm{mm}\) and \(24\,\mathrm{mm}\) height increments, cutting was still completed, but the increased board height caused larger vertical forces during knife raising. 
These larger forces degraded horizontal tracking over repeated slices, resulting in noticeably thinner slices.
In comparison, the proposed VMC maintained lower average and peak contact forces than the hybrid force/position controller under large table height variation, demonstrating robustness to environmental variation.


\subsection{Portability}

\begin{figure}[htbp]
	\centering
	\begin{subfigure}[h]{.8\linewidth}
		\includegraphics[width=\columnwidth]{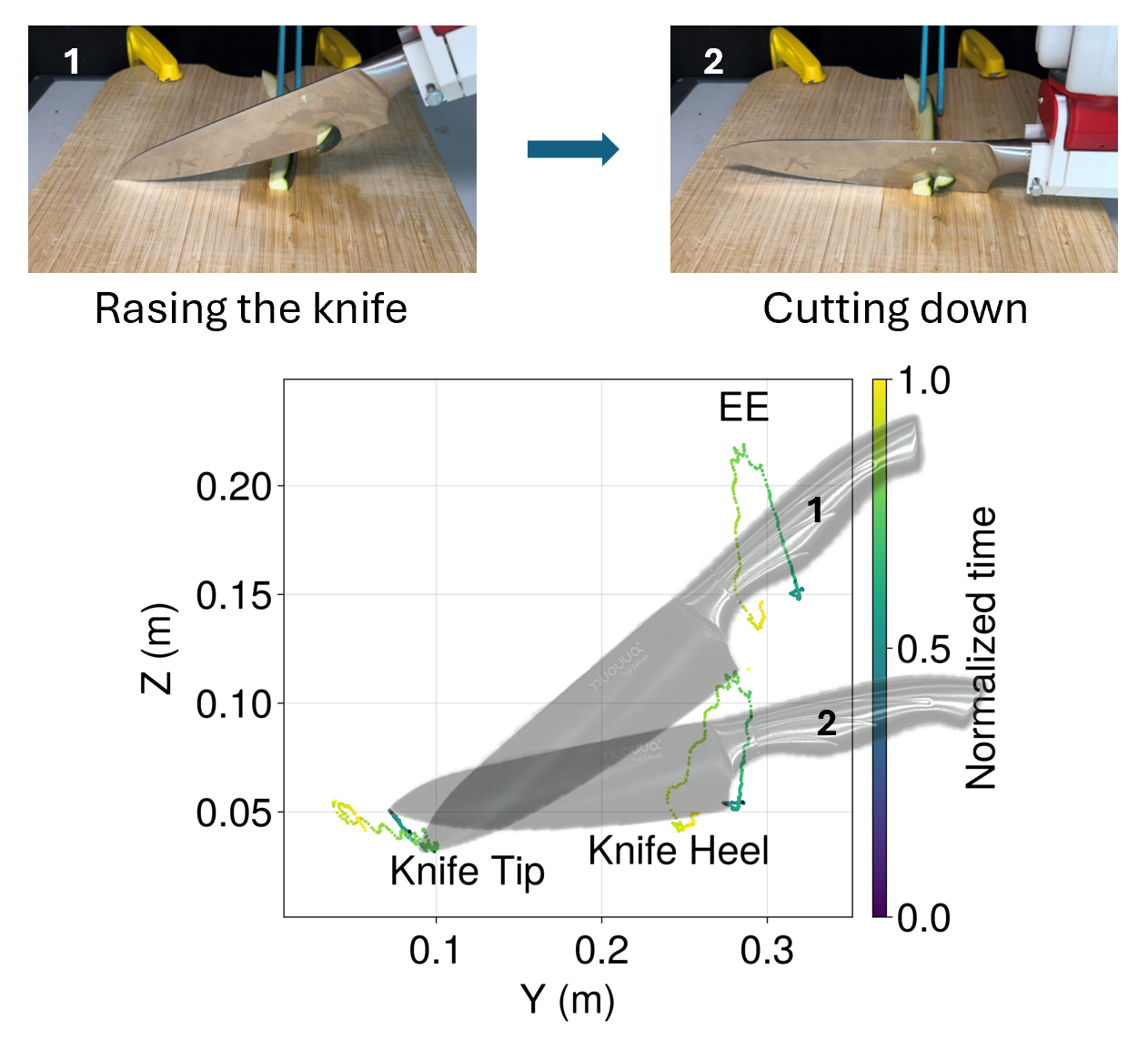}
		\caption{Snapshots of Sciurus17 rocking with a knife.}
		\label{fig:exp_1_2_rocking_snapshots}
	\end{subfigure}

	\begin{subfigure}[t]{.99\linewidth}
		\includegraphics[width=\columnwidth]{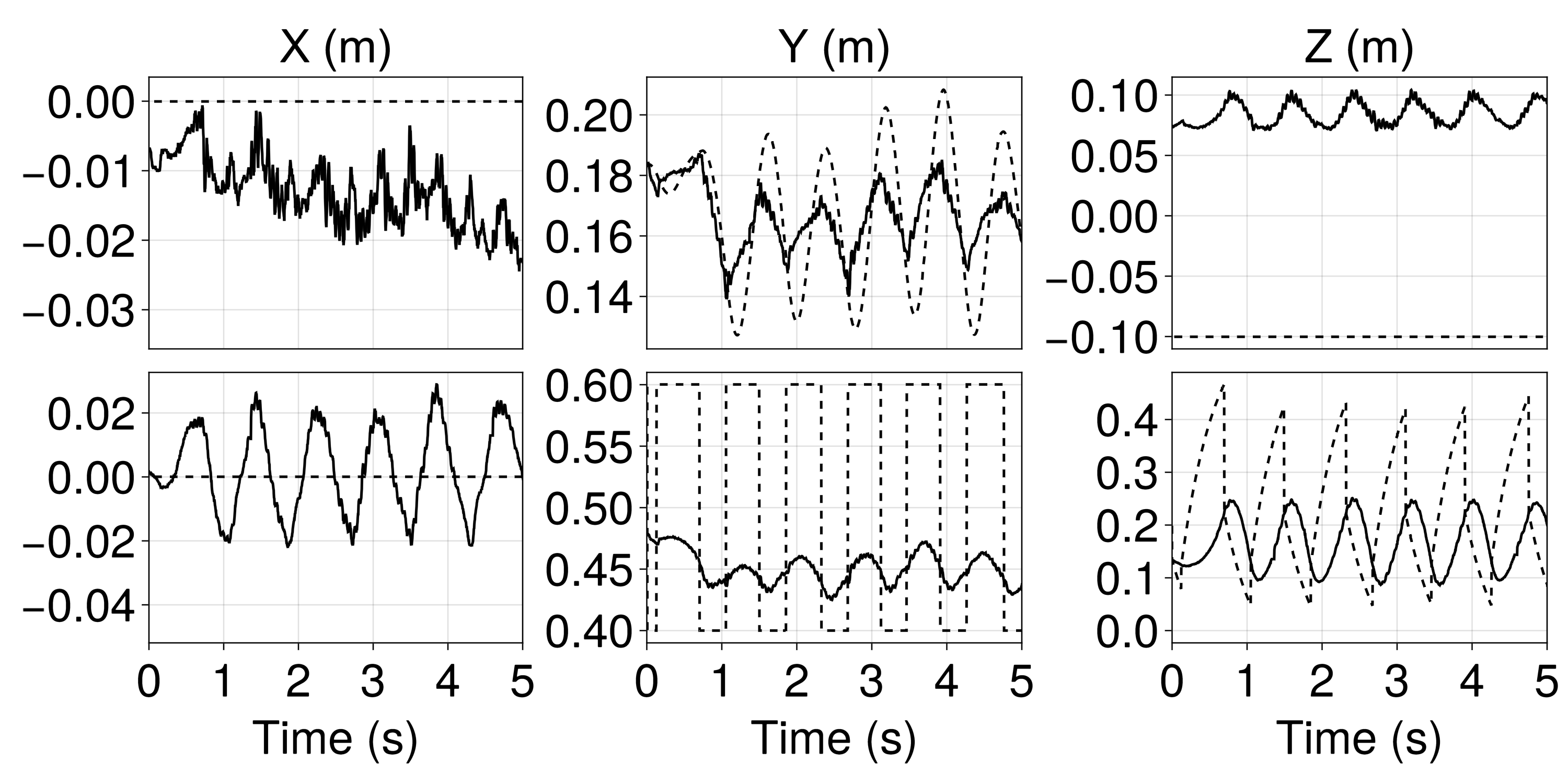}
		\caption{Top: solid - $\pmb{p}_1$, dashed - $\pmb{p}_a$. 
			Bottom: solid - $\pmb{p}_2$, dashed - $\pmb{p}_b$.}
		\label{fig:rocking_1_2_p1_q1}
	\end{subfigure}
	\caption{Rocking motion Sciurus17. The virtual model controller is
		unchanged but its parameters have been adjusted.}
	\label{fig:exp_1_2_sciurus_data}
\end{figure}

The proposed virtual mechanism is platform independent. 
We demonstrate this by implementing the same controller on a humanoid robot from RT Corp, Sciurus17.
Unlike the Franka platform, Sciurus17 has a lower payload capacity, slower communication speed and uses current control as a proxy of torque control. 
These differences can be managed by re-tuning the parameters of the controller to ensure smaller forces and slower motions. 
The cutting plane was kept fixed, and food items were manually fed to the knife to reduce system burden.
Fig. \ref{fig:exp_1_2_sciurus_data} shows that the inherent limitations of the robotic platform led to reduced performance. However, successful rocking motion is achieved. 

\section{Conclusion and future works} 
We present a VMC-based controller for robust, periodic rock-chopping that generates a limit cycle through switched mechanisms. The controlled rocking motion is analyzed through energy considerations, and its periodicity is numerically certified via contraction of the associated Poincaré map.
The controller was validated both in simulations and experimentally, focusing on parameter sensitivity and robustness to food and environmental uncertainties.
Performance has been illustrated by cutting several vegetables, for several requirements of thicknesses and cut frequencies.
Robustness has been illustrated by using two different knives and several cutting board heights. 
The parameter sensitivity analysis further shows that the mechanism tolerates  variations in geometric placement and all other parameters in the controller.

Figs. \ref{fig:exp_2_knife_result} and \ref{fig:exp_2_height_result} clarify the difference between the proposed virtual model controller, Cartesian impedance tracking, and the hybrid force/position baseline. 
The impedance and hybrid baselines remain tied to a nominal trajectory, making them sensitive to mismatch in knife geometry or cutting-board height. 
In addition, a pre-planned trajectory would typically require precise knowledge of knife geometry and high-gains for accurate tracking.
In contrast, the rocking motion established by the virtual model controller is not pre-programmed but shaped by the interaction between robot, controller and environment, therefore robustly adapts to a wide range of settings.
The experimental evaluation of the impedance controller shows the limits of shaping the end-effector impedance without explicitly taking into account the geometry of the task. In contrast, the design of the virtual model controller starts from the task geometry to structure the compliance of the virtual mechanisms in a task-relevant way. This leads to more flexible tuning and increased robustness. For instance, the maximum force $\sigma_1$ affects directly the contact force with the cutting board. Likewise, the maximum force $\sigma_{\mathrm{ori}}$ penalizes deviations from the slicing plane.
Experiment videos are available in Extension 1.

Successful operation depends on reasonable parameter selection and approximate task-level initialization. The controller requires an approximate object location and knife pointing direction relative to the end effector, but does not require precise knife geometry, exact cutting-board height, or exact virtual point placement as demonstrated in the sensitivity analysis.
These features distinguish our approach from \cite{2023YanBin}, which considers rolling knife motion but formulates cutting as a staged controller with different controller modes for pressing, touching, and slicing. 
That formulation relies on an explicit knife-geometry model to plan contact-point transitions and compute the corresponding Jacobians. 
In contrast, our VMC controller generates repeated, robust and fast rock-chop motion through the interaction between the robot, virtual mechanism, knife, and environment, without tracking a preplanned rolling trajectory or switching among controller types. 
This makes the controller suitable for repeated cutting under environment uncertainty.

\begin{figure}[htbp]
	\centering
	\begin{subfigure}[t]{.49\linewidth}
		\includegraphics[width=\columnwidth]{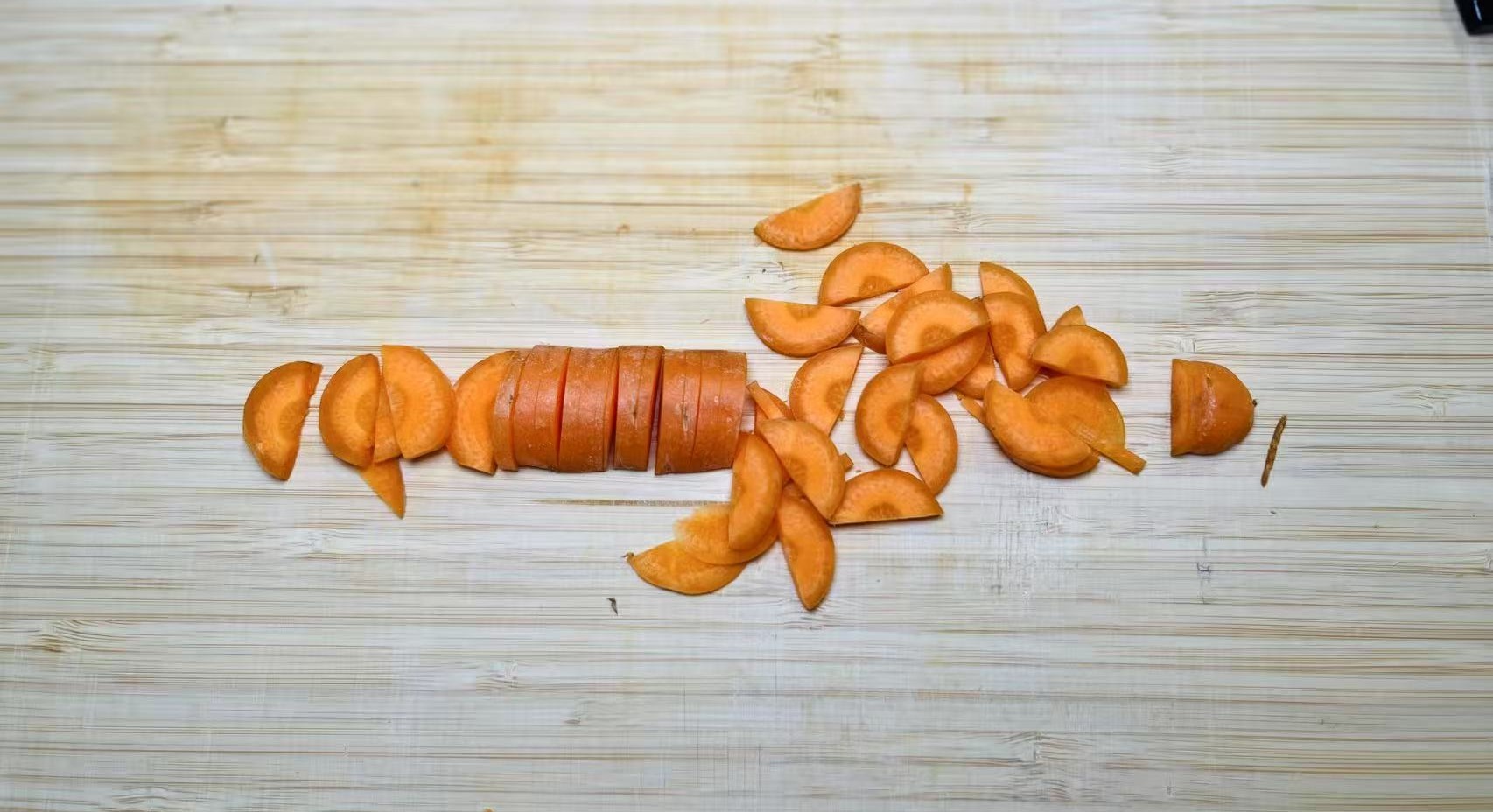}
		\caption{Carrot cut result.}
	\end{subfigure}
	\begin{subfigure}[t]{.48\linewidth}
		\includegraphics[width=\columnwidth]{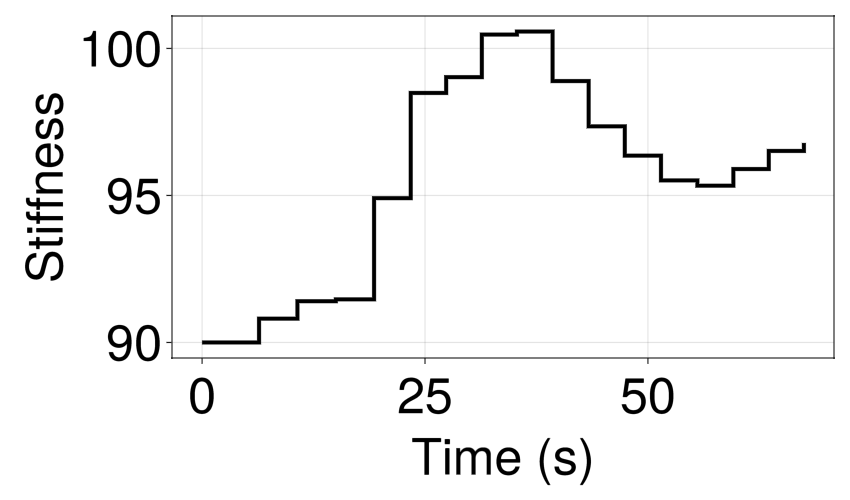}
		\caption{Stiffness $k_2$.}
		\label{fig:adaptation_stiffness_b}
	\end{subfigure}
	
	\caption{Adaptation of spring stiffness for cutting a carrot.}
	\label{fig:adaptation_stiffness}
\end{figure}

The proposed controller addresses the control layer and can be integrated into a full cutting pipeline.
This involves autonomous pick and place of the food, its orientation on the cutting board, and the ability to hold the food in place during cutting. The robot should also be able to recognize different foods and knives, and to measure in real-time the result of its action, during and after cutting. 
At the level of food-knife interaction, a key future direction is the systematic exploration of adaptation/learning of the control parameters, for stronger performance. The geometric configuration of the virtual model controller serves as the primary driver of the robot's behavior, while parameters such as stiffness and damping are used for fine-tuning. This provides a fundamental advantage for tuning, as the geometry defines the core motion while the control parameters refine it. The objective is to exploit this structural advantage in combination with external sensing, such as vision and tactile feedback, to enable real-time parameter adaptation. For example, the stiffness $k_2$ can be adjusted in real-time to achieve successful food separation while moderating the force exerted by the knife on food and board.
The short experimental study proposed below shows encouraging preliminary results. Every six cutting cycles we increase 
$k_2$ if the knife does not achieve separation of food and
we reduce $k_2$ if the knife's handle overshoot is too large 
(i.e. if the knife handle moves past the critical reference). 
This leads to the adaptation law 
\[
k_2^{(n+1)} = k_2^{(n)} + \alpha\, e,
\]
where $e$ is the error signal capturing the mismatch between desired and achieved vertical position of the knife, and $\alpha$ is the adaptation gain. Results are summarized in Fig. \ref{fig:adaptation_stiffness}.
$k_2$ starts from a low value, which is sufficient to cut through a carrot. From time $t=20$s, the carrot presents a harder consistency which requires adaptation. Successful cutting is restored after time $t=35s$ through adaptation of $k_2$.

\begin{table*}[t]
\centering 
\caption{\textbf{Control parameters and sensitivity analysis.} Units: stiffness $k$ [N/m], maximum force $\sigma$ [N], damping $c$ [N·s/m], position $\pmb{r}$ [m], mass $m$ [kg], $\delta$ [m], $\tau$ [s]. The mass–spring–damper system ($m_b, k_b, c_b$) generates smooth transitions between reference points. Parameters are selected to ensure overdamped behavior, and the time constant $\tau$ of the slower pole is reported as an intuitive measure of the transition dynamics. \emph{Consistent parameters across all experiments in simulation and real-world}: masses for simulating the sliding motion ($m_a=m_{\text{ori}}=0.1$), $\pmb{p}_{a}(z)=-0.1$. 
		\emph{Consistent parameters across all experiments on Franka in simulation and real-world}: $\sigma_1=10, c_1=1, \sigma_{\text{ori}}=50, c_{\text{ori}}=10$, $\pmb{r}_{2,1}\, (y,z)= 0.43,  -0.1$, $\pmb{p}_{1}=(0.0, 0.2, 0.05)$, $\pmb{p}_{2}=(0.0, -0.2, 0.05)$, $\pmb{p}_{3}=(0.0, 0.5, 0.0)$, $\pmb{p}_{4}=(0.0,-0.5,0.0)$, $\pmb{p}_{5}=(0.0, 0.0, 0.5)$, and $\pmb{p}_{6}=(0.0, 0.0, -0.5)$. \vspace{2mm}}
        \label{tab:sensitivity}
        
\begin{subtable}[t]{\textwidth}
    \centering
    \caption{\textbf{Simulation parameters for limit-cycle and sensitivity analysis.}
    \label{tab:sensitivity_parameters}
The parameters match those used in the robustness experiments in Table \ref{tab:control_parameters}(c), except for $k_1$, $\pmb{r}_{2,2}$ and $\tau$, which were adjusted to account for differences between the simulated and real robot dynamics.  Robot: Franka.
}
    {
    \small
    \centering
        \begin{tabular}{|p{3.8cm}| c|c|c|c|c|c|c|c|c|}
            \hline
            Simulation & $k_1$ & $k_2$ & $\sigma_2$ & $c_2$ & $k_{\text{ori}}$ & $\pmb{r}_{2,2}\, (y,z)$ & $\delta_1$ & $\delta_2$ & $\tau$ \\ \hline
              & 60 & 130 & 25 & 2 & 800 & 0.53, 0.35 & 0.01 & 0.2  & 0.2 \\ \hline
        \end{tabular} \vspace{2mm}
    }
\end{subtable}
\begin{subtable}[t]{\textwidth}
	\centering
    \caption{Sensitivity analysis of limit-cycle stability and cutting performance under controller parameter variations. A failure indicates that the modified parameter did not produce a valid periodic cutting motion.}
    \label{tab:sensitivity_results}

    \begin{tabular}{c c c c c c c}
    \hline
    Parameter & Variation & $\max_i|\lambda_i|$ & Stability & 
    $F_{\mathrm{avg}}$ [N] & 
    $F_{\mathrm{peak}}$ [N] & 
    $f_{\mathrm{cut}}$ [Hz] \\
    \hline
    Nominal & -- & 0.93 & Stable & 40.2 & 48.7 & 0.48 \\ 
    $k_1$ & $-20\%$ & 0.86 & Stable & 39.8 & 48.3 & 0.51 \\
    $k_1$ & $+20\%$ & 0.90 & Stable & 39.8 & 48.6 & 0.46 \\
    $c_1$ & $-20\%$ & 0.91 & Stable & 40.2 & 48.7 & 0.48 \\
    $c_1$ & $+20\%$ & 0.91 & Stable & 40.2 & 48.7 & 0.48 \\
    $\sigma_1$ & $-20\%$ & 0.87 & Stable & 39.8 & 48.4 & 0.51 \\
    $\sigma_1$ & $+20\%$ & 0.90 & Stable & 39.9 & 48.7 & 0.46 \\
    $\pmb{p}_{1}$ & $-20\%$ & 0.93 & Stable & 39.3 & 49.5 & 0.68 \\ 
    $\pmb{p}_{1}$ & $+20\%$ & 0.89 & Stable & 38.8 & 47.8 & 0.28 \\ 
   $\pmb{p}_{a}(z)$ & $-20\%$ & 0.95 & Stable & 39.6 & 48.5 & 0.50 \\ 
   $\pmb{p}_{a}(z)$ & $+20\%$ & 0.92 & Stable & 40.1 & 48.9 & 0.47 \\ 
    $k_2$ & $-20\%$ & 0.96 & Stable & 35.8 & 42.7 & 0.31 \\
    $k_2$ & $+20\%$ & 0.91 & Stable & 42.4 & 53.2 & 0.62 \\
    $c_2$ & $-20\%$ & 0.90 & Stable & 40.2 & 48.9 & 0.47 \\
    $c_2$ & $+20\%$ & 0.92 & Stable & 39.4 & 48.1 & 0.49 \\
    $\sigma_2$ & $-20\%$ & 0.88 & Stable & 36.8 & 42.6 & 0.32 \\
    $\sigma_2$ & $+20\%$ & 0.90 & Stable & 39.7 & 50.1 & 0.63 \\
   $\pmb{p}_{2}$ & $-20\%$ & 0.82 & Stable & 37.0 & 44.7 & 0.21 \\ 
   $\pmb{p}_{2}$ & $+20\%$ & 0.84 & Stable & 38.8 & 49.5 & 0.68 \\ 
    $\pmb{r}_{2,1}(z)$ & $-20\%$ & 0.95 & Stable & 38.2 & 46.0 & 0.37 \\
    $\pmb{r}_{2,1}(z)$ & $+20\%$ & 0.91 & Stable & 41.2 & 51.3 & 0.61 \\
    $\pmb{r}_{2,2}(z)$ & $-20\%$ & - & Failure & - & - & - \\
    $\pmb{r}_{2,2}(z)$ & $+20\%$ & 0.51 & Stable & 43.5 & 55.6 & 0.11 \\
    $\pmb{r}_{2,2}(y)$ & $-20\%$ & 0.89 & Stable & 37.8 & 46.5 & 0.45 \\
    $\pmb{r}_{2,2}(y)$ & $+20\%$ & 0.88 & Stable & 40.0 & 49.1 & 0.54 \\
    $\delta_1$ & $-20\%$ & 0.88 & Stable & 39.1 & 48.2 & 0.46 \\
    $\delta_1$ & $+20\%$ & 0.88 & Stable & 40.2 & 48.6 & 0.51 \\
    $\delta_2$ & $-20\%$ & 0.79 & Stable & 43.9 & 48.9 & 0.24 \\
    $\delta_2$ & $+20\%$ & 0.76 & Stable & 30.1 & 39.6 & 1.00 \\
   $\tau$ & $-20\%$ & 0.96 & Stable & 39.9 & 49.2 & 0.54 \\ 
   $\tau$ & $+20\%$ & 0.93 & Stable & 36.5 & 45.7 & 0.44 \\ 
   $\pmb{p}_{3}-\pmb{p}_{6}$ & $-20\%$ & 0.90 & Stable & 38.0 & 47.0 & 0.49 \\
   $\pmb{p}_{3}-\pmb{p}_{6}$ & $+20\%$ & 0.97 & Stable & 40.2 & 48.7 & 0.48 \\ 
    $k_{ori}$ & $-20\%$ & 0.89 & Stable & 40.2 & 48.7 & 0.48 \\
    $k_{ori}$ & $+20\%$ & 0.92 & Stable & 40.2 & 48.7 & 0.48 \\
    $c_{ori}$ & $-20\%$ & 0.90 & Stable & 39.5 & 48.6 & 0.48 \\
    $c_{ori}$ & $+20\%$ & 0.91 & Stable & 40.2 & 48.7 & 0.48 \\
    $\sigma_{ori}$ & $-20\%$ & 0.91 & Stable & 39.6 & 48.6 & 0.48 \\
    $\sigma_{ori}$ & $+20\%$ & 0.91 & Stable & 40.2 & 48.7 & 0.48 \\
    \hline
    \end{tabular}

\end{subtable}
\end{table*}

\begin{table*}[htbp]
	\centering
	\caption{\textbf{Control parameters for each experiment. }
		Units: stiffness $k$ [N/m], maximum force $\sigma$ [N], damping $c$ [N·s/m], position $\pmb{r}$ [m], mass $m$ [kg], $\delta$ [m], $\tau$ [s].
		The mass–spring–damper system ($m_b, k_b, c_b$) generates smooth transitions between reference points. Parameters are selected to ensure overdamped behavior, and the time constant $\tau$ of the slower pole is reported as an intuitive measure of the transition dynamics.
		\emph{Consistent parameters across all experiments in simulation and real-world}: masses for simulating the sliding motion ($m_a=m_{\text{ori}}=0.1$), $\pmb{p}_{a}(z)=-0.1$. 
		\emph{Consistent parameters across all experiments on Franka}: $\sigma_1=10, c_1=1, \sigma_{\text{ori}}=50, c_{\text{ori}}=10$, $\pmb{r}_{2,1}\, (y,z)= 0.43,  -0.1$, $\pmb{p}_{1}=(0.0, 0.2, 0.05)$, $\pmb{p}_{2}=(0.0, -0.2, 0.05)$, $\pmb{p}_{3}=(0.0, 0.5, 0.0)$, $\pmb{p}_{4}=(0.0,-0.5,0.0)$, $\pmb{p}_{5}=(0.0, 0.0, 0.5)$, and $\pmb{p}_{6}=(0.0, 0.0, -0.5)$. \vspace{2mm}
	}
	\label{tab:control_parameters}
	\begin{subtable}[t]{\textwidth}
		\centering
		\caption{\textbf{Uniform control parameters and parameters tuned to reduce force for individual vegetables. }
			The most sensitive parameters here are the stiffness values ($k_1$, $k_2$) governing contact and cutting. 
			$m_b, k_b, c_b$ regulate the transition between reference points $\pmb{r}_{2,1}$ and $\pmb{r}_{2,2}$ to maintain similar cutting frequency when $k_2$ is reduced. $\pmb{r}_{2,2}$ and $\delta_2$ are adjusted according to object height.  Robot: Franka.}
		{\small
			\begin{tabular}{|p{3.2cm}|c|c|c|c|c|c|c|c|c|}
				\hline
				Exp & $k_1$  & $k_2$ & $\sigma_2$ & $c_2$ & $k_{\text{ori}}$ & $\pmb{r}_{2,2}\, (y,z)$ & $\delta_1$ & $\delta_2$ & $\tau$  \\ \hline
				Uniform parameters & 25 & 150 & 25 & 2 & 1200 & 0.58, 0.4 & 0.02 & 0.15 & 0.5 \\ \hline
				Spring onion  & 10 & 65  & 25 & 2 & 1200 & 0.58, 0.4 & 0.02 & 0.2  & 0.05 \\ \hline
				Cucumber      & 10 & 75  & 25 & 2 & 1200 & 0.83, 0.5 & 0.02 & 0.25 & 0.05  \\ \hline
				Courgette     & 10 & 105 & 25 & 2 & 1200 & 0.83, 0.5 & 0.02 & 0.25 & 0.05  \\ \hline
				Carrot        & 20 & 105 & 25 & 2 & 1200 & 0.83, 0.5 & 0.02 & 0.25 & 0.05  \\ \hline
				Potato        & 20 & 105 & 25 & 2 & 1200 & 0.83, 0.5 & 0.02 & 0.25 & 0.05  \\ \hline
			\end{tabular}
		}
	\end{subtable}
	
	\vspace{4mm}
	
	\begin{subtable}[t]{\textwidth}
		\centering
		\caption{\textbf{Different cutting frequencies. }
			The primary tuning parameter is $k_2$, increased for faster cuts. 
			$m_b$, $k_b$, and $c_b$ are adjusted to switch faster, while $k_1$ is increased to ensure stable contact during rapid motion. $\delta_1$ is increased for faster cuts to accommodate higher knife velocity and avoid overshoot during motion reversal. Robot: Franka.}
		{\small
			\begin{tabular}{|p{3cm}|c|c|c|c|c|c|c|c|c|}
				\hline
				Exp & $k_1$ & $k_2$ & $\sigma_2$ & $c_2$ & $k_{\text{ori}}$ & $\pmb{r}_{2,2}\, (y,z)$ & $\delta_1$ & $\delta_2$ & $\tau$ \\ \hline
				Fast cut (0.94 cut/s) & 45 & 200 & 30 & 2 & 800  & 0.73, 0.35 & 0.035 & 0.145 & 0.015  \\ \hline
				Slow cut (0.51 cut/s) & 25 & 120 & 25 & 2 & 800  & 0.58, 0.4  & 0.02  & 0.15  & 0.5 \\ \hline
			\end{tabular}
		}
	\end{subtable}
	
	\vspace{4mm}
	
	\begin{subtable}[t]{\textwidth}
		\centering
		\caption{\textbf{Robustness to environment variations. } \label{tab:robust_experiment}
			Identical mechanism and control parameters are used across all robustness experiments. Robot: Franka.}
		{\small
			\begin{tabular}{|p{3.8cm}|c|c|c|c|c|c|c|c|c|}
				\hline
				Exp & $k_1$ & $k_2$ & $\sigma_2$ & $c_2$ & $k_{\text{ori}}$ & $\pmb{r}_{2,2}\, (y,z)$ & $\delta_1$ & $\delta_2$ & $\tau$ \\ \hline
				Robustness & 30 & 130 & 25 & 2 & 800 & 0.63, 0.4 & 0.01 & 0.2  & 0.1 \\ \hline
			\end{tabular}
		}
	\end{subtable}
	
	\vspace{4mm}
	
	\begin{subtable}[t]{\textwidth}
		\centering
		\caption{\textbf{Robustness to environment variations. }
			Impedance gains used for the trajectory-tracking baseline. Translational stiffness is given along the Cartesian axes, and rotational stiffness about the corresponding axes. Robot: Franka.}
		{\small
			\begin{tabular}{|l|ccc|ccc|}
				\hline
				& \multicolumn{3}{c|}{Translational stiffness} & \multicolumn{3}{c|}{Rotational stiffness} \\
				& $K_x$ & $K_y$ & $K_z$ & $K_{r_x}$ & $K_{r_y}$ & $K_{r_z}$ \\
				\hline
				High gain & 1500 & 7000 & 7000 & 300 & 170 & 170 \\
				\hline
				Low gain  &  750 & 3500 & 3500 & 150 &  85 &  85 \\
				\hline
                Hybrid force/position & 1500 & 7000 & 3500 & 300 & 170 & 170 \\
				\hline
			\end{tabular}
		}
	\end{subtable}
	
	\vspace{4mm}
	
	\begin{subtable}[t]{\textwidth}
		\centering
		\caption{\textbf{Platform independence. }
			Parameters are adjusted to account for differences in actuation capability and communication frequency between platforms. Stiffness and max forces are reduced. Points $\pmb{p}_{1}$--$\pmb{p}_{6}$ are scaled toward the robot to further reduce the required force. Robot: Sciurus.}
		{\small
			\begin{tabular}{|c|c|c|c|c|c|c|c|c|c|c|c|c|c|}
				\hline
				$k_1$ & $\sigma_1$ & $c_1$ & $k_2$ & $\sigma_2$ & $c_2$ & $k_{\text{ori}}$ & $\sigma_{\text{ori}}$ & $c_{\text{ori}}$ & $\pmb{r}_{2,1}\, (y,z)$ & $\pmb{r}_{2,2}\, (y,z)$ & $\delta_1$ & $\delta_2$ & $\tau$ \\ \hline
				10 & 2 & 0.3 & 20 & 10 & 0.1 & 30  & 10  & 1 & 0.4, -0.1 & 0.6, 0.6 & 0.047 & 0.38  & 0.45 \\ \hline
			\end{tabular}
            \vspace*{2mm}
            
            \begin{tabular}{|c|c|c|c|c|c|}
				\hline
				$\pmb{p}_{1}$ & $\pmb{p}_{2}$ & $\pmb{p}_{3}$ & $\pmb{p}_{4}$ & $\pmb{p}_{5}$ & $\pmb{p}_{6}$  \\ \hline
				(0.0, 0.15, 0.05) & (0.0, -0.15, 0.05) & (0.0, 0.2, 0.0) & (0.0, -0.2, 0.0) & (0.0, 0.0, 0.2) & (0.0, 0.0, -0.2)  \\ \hline
			\end{tabular}
		}
	\end{subtable}

\end{table*}

\begin{acks}
	Yi Zhang is supported by the Engineering and Physical Sciences Research Council, AgriFoRwArdS Center for Doctoral Training [EP/S023917/1] and RT Corp (Japan).
    \vspace{0.55cm}
\end{acks}

\bibliographystyle{SageH}

\end{document}